\RequirePackage{fix-cm}
\documentclass[smallextended]{svjour3}       % onecolumn (second format)
\smartqed  % flush right qed marks, e.g., at end of proof
\usepackage[table]{xcolor}

\usepackage{multirow}
\usepackage{times}

\usepackage{hyperref}

\usepackage{color}
\usepackage{algorithm}
\usepackage{algpseudocode}
\usepackage{listings}
\usepackage{courier}
\usepackage{soul}
\usepackage{amsmath,amsfonts,amssymb}

\usepackage{mathtools}
\usepackage{subfigure}% Support for small, `sub' figures and tables
\usepackage{wrapfig}
\usepackage{hyperref}
\usepackage{multirow}
\usepackage{graphicx}
\usepackage{soul}
\usepackage[flushleft]{threeparttable}
\usepackage{enumerate}
\usepackage{algorithmicx}
\usepackage{algpseudocode}
\usepackage{epstopdf}
\usepackage{booktabs}

\definecolor{red}{rgb}{1.00,0.00,0.00}
\definecolor{blue}{rgb}{0.00,0.00,1.00}
\definecolor{green}{rgb}{0.2,0.70,0.2}
\definecolor{yellow}{rgb}{0.5,0.5,0.0}
\definecolor{white}{rgb}{1,1,1}

\begin{document}

\title{ A Modular Framework to Generate Robust Biped Locomotion: From Planning to Control
%|	\thanks{This research is supported by Portuguese National Funds through Foundation for Science and Technology (FCT) through FCT scholarship SFRH/BD/118438/2016 and also in the context of the project UIDB/00127/2020.}
 }

\institute{Mohammadreza Kasaei, Nuno Lau, Artur Pereira \at
              IEETA / DETI University of Aveiro 3810-193 Aveiro, Portugal  \\
              Tel.: +351-234-370-500\\
              Fax: +351-234-370-545\\
              \email{\{Mohammadreza, nunolau, artur\}@ua.pt} 
           \and
           Ali Ahmadi \at
              \email{Ahmadi.bps@gmail.com}
}

\date{Received: date / Accepted: date}
\author{Mohammadreza Kasaei\and Ali Ahmadi\and Nuno Lau \and Artur Pereira }

\maketitle

\begin{abstract}
Biped robots are inherently unstable because of their complex kinematics as well as dynamics. Despite many research efforts in developing biped locomotion, the performance of biped locomotion is still far from the expectations. This paper proposes a model-based framework to generate stable biped locomotion. The core of this framework is an abstract dynamics model which is composed of three masses to consider the dynamics of stance leg, torso, and swing leg for minimizing the tracking problems. According to this dynamics model, we propose a modular walking reference trajectories planner which takes into account obstacles to plan all the references. Moreover, this dynamics model is used to formulate the controller as a Model Predictive Control~(MPC) scheme which can consider some constraints in the states of the system, inputs, outputs, and also mixed input-output. The performance and the robustness of the proposed framework are validated by performing several numerical simulations using~\mbox{MATLAB}. Moreover, the framework is deployed on a simulated torque-controlled humanoid to verify its performance and robustness. The simulation results show that the proposed framework is capable of generating biped locomotion robustly.

\end{abstract}

\section{Introduction}
\label{sec:introduction}
Humanoid robots are more adapted to our real environment for helping us to perform our daily-life tasks. Developing a robust walking framework for humanoid robots has been researched for decades and it is still a challenging problem in the robotics community. The complexity of this problem derives from several different aspects like considering an accurate hybrid dynamics model, designing appropriate controllers, and formulating suitable reference trajectory planners. The wide range of applications for these type of robots motivates researchers to tackle such a complex problem using different approaches which can be generally classified into two main classes: \textit{model-free}~\cite{santos2017biped,tran2018humanoid} and \textit{model-based}~\cite{chang2015humanoid,faraji20173lp,kasaei2019fast}. %These approaches will be briefly reviewed in the reset this section.

Model-free approaches try to generate walking patterns by generating rhythmic patterns for each limb and without considering any dynamics model of the system. Some of them are inspired from neuro-physiological studies on humans and animals. These studies showed that periodic locomotion such as walking and running are generated by a set of oscillators at spinal cord which are connected together in a specific arrangement. Each oscillator has a set of parameters that are generally tuned by some trail-intensive method like trials and error, machine learning~(ML) algorithms or both of them~\cite{wang2018matsuoka,saputra2019layered}. Some other approaches in this class are designed based on learning from scratch and mostly are based on Reinforcement Learning~(RL) algorithms~\cite{abreu2019learning,li2013humanoids} which need many samples to be able to generate walking that takes a considerable amount of time. Unlike the model-free approaches, the fundamental core of the model-based approach is a dynamics model of the robot. The main idea of these approaches is modeling the overall behavior of the system in an abstract manner, then planners and controllers will be formulated according to this model. The first question in designing a model-based walking is \textit{how accurate should the model be?}. To answer this question, two points of view exist: (i)~using a whole-body dynamics model, (ii)~using an abstract model. We believe that for designing a dynamics model of a system, a trade-off between accuracy and simplicity should be taken into account. Although the whole body dynamics model is more accurate than an abstract model, it is not only platform-dependent but also needs a high computational effort. The main idea behind using an abstract model is fading the complexity of the control system and organizing the control system as a hierarchy. In hierarchical control approaches, a simplified dynamics model is used to abstract the overall behaviors of the system, and then these behaviors will be converted to individual actuator inputs using a detailed full-body inverse dynamics~\cite{faraji20173lp}. In such approaches, the overall performance of the system depends on the matching between the abstract model and the exact model.

In this paper, we proposed a modular model-based walking framework capable of generating robust locomotion for humanoid robots. In this framework, the overall dynamics of the robot is modeled using a three-mass model which takes into account dynamics of the legs and the torso. According to this model, the problem of dynamics locomotion will be formulated as a linear MPC to predict the behavior of the system over a prediction horizon and to determine the optimum control inputs. Additionally, the process of generating walking reference trajectories will be decomposed into three levels including (i) path  and footstep planning, (ii) planning ZMP, hip and swing reference trajectories and (iii) planning a set of reference trajectories for the controller. Moreover, the proposed framework has a hierarchical structure which is composed of several reusable blocks. The designed architecture reduces the complexity of the framework and can be used to advancements in research and development. Besides, the results of several simulations will be presented to show the performance and the robustness of the proposed walking scheme.

Our contribution is twofold. First, the development of a modular framework that reduces the complexity and increase the flexibility of generating robust locomotion. Second, formulating the walking controller as a Linear MPC based on three-mass model.
The remainder of this paper is structured as follows: Section~\ref{sec:related_work} gives an overview of the related work. The overall architecture of the framework will be presented in Section~\ref{sec:arct}. Later, in Section~\ref{sec:ReferenceTrajectories}, the reference trajectory planners will be presented. Afterwards, the three-mass dynamics model will be reviewed briefly in Section~\ref{sec:DynamicsModel} and then we will explain how this model will be used to design a walking controller based on MPC scheme in Section~\ref{sec:MPC}.  In Section~\ref{sec:simulation}, a set of simulation scenarios will be designed and performed to examine the tracking performance and robustness of the proposed controller. Afterwards, in Section~\ref{sec:Discussion}, a baseline framework based on LIPM will be used to compare and highlight the effectiveness of the proposed framework. Finally, conclusions and future research will be presented in Section~\ref{sec:CONCLUSION}.

\begin{figure}[!t]
	\centering
	\begin{tabular}	{c c}	
		\includegraphics[width = 0.45\columnwidth, trim= 13.0cm 5cm 12cm 5cm,clip] {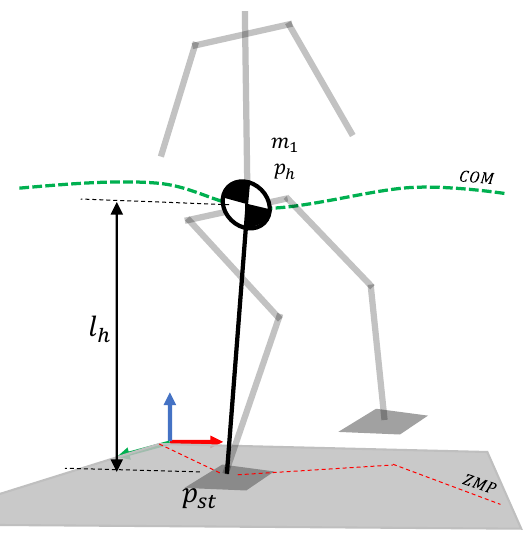} & \hspace{-5mm}
		\includegraphics[width = 0.45\columnwidth, trim= 13.0cm 5cm 12cm 5cm,clip]{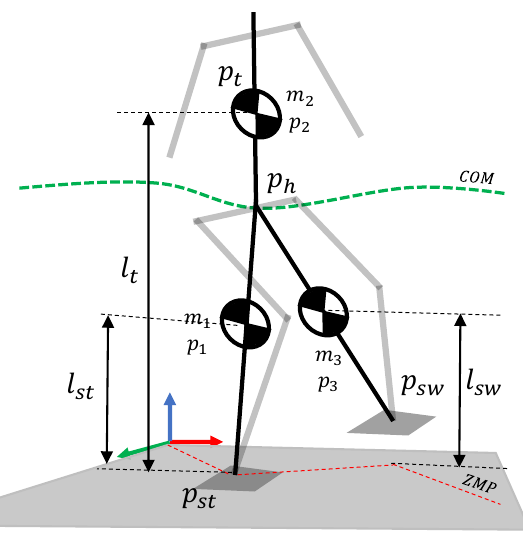} \\
		\textbf{(a)} Linear Inverted Pendulum & \textbf{(b)} Three-mass model
	\end{tabular}
	\vspace{-0mm}
	\caption{ Abstract dynamics models: \textbf{(a)} represents the Linear Inverted Pendulum Model~(LIPM) which abstracts the overall dynamics of a humanoid robot into a single mass with restricted vertical motion; \textbf{(b)} represents the three-mass model which takes into account the dynamics of the legs and the torso with restricted vertical motions.}
	%	\vspace{-5mm}
	\label{fig:dynamicsmodel}
\end{figure}

\section {Related Work}
\label{sec:related_work}

Several simplified dynamics models have been proposed that abstract the overall dynamics of humanoid robots. In this section, some of these models will be reviewed briefly and then some recent works that used these models to generate robust locomotion or to develop push recovery strategies will be presented.

\subsection{Dynamics Models}
Linear Inverted Pendulum Model~(LIPM) is one of the common dynamics models in the literature. The popularity of this model originates from its linearity and simplicity and from its ability to generate a feasible, fast, and efficient trajectory of the Center of Mass~(COM)~\cite{kajita2001real}. This model describes the dynamics of a humanoid robot just by considering a single mass that is connected to the ground via a mass-less rod~(see Fig.~\ref{fig:dynamicsmodel}\textbf{(a)}). In this model, the single mass is assumed to move along a horizontal plane and based on this assumption, the motion equations in sagittal and frontal planes are decoupled and independent. Several studies used this model to develop an online walking generator based on optimal control~\cite{KASAEIIROS} and also linear MPC~\cite{herdt2010walking,brasseur2015robust,mirjalili2018whole}. Several extensions of LIPM have been proposed to increase the accuracy of this model while keeping the simplicity level~\cite{shimmyo2013biped,faraji20173lp,kasaei2017reliable,luo2016humanoid}. The three-mass model is one of the extended versions of LIPM~\cite{takenaka2009real,sato2010real,shimmyo2013biped}. As it is shown in Fig.~\ref{fig:dynamicsmodel}\textbf{(b)}, this model considers the masses of legs and the torso to increase the accuracy of the model. It should be mentioned that to keep the model linear, the vertical motions of the masses are considered to be smooth and the vertical accelerations are neglected.

Kajita et al.~\cite{kajita2001real} introduced the Three Dimensional Linear Inverted Pendulum Model~(3D-LIPM) and explained how this model can be used to generate walking. Afterwards, in~\cite{kajita2003biped}, they used the concept of ZMP to develop a control framework based on preview control to generate stable locomotion. The performance of their framework has been validated using several simulations.

Albert and Gerth~\cite{albert2003analytic} extended LIPM by considering the mass of the swing leg to improve gait stability and they named it Two Masses Inverted Pendulum
Model (TMIPM). In their approach, the swing leg and the ZMP trajectories should be generated firstly, then the trajectory of the COM will be generated by solving a linear differential equation. They proposed an extended version of TMIPM which considers the mass of the thigh, the shank, and the foot of the swinging leg which has been called Multiple Masses Inverted Pendulum Model~(MMIPM). TMIPM and MMIPM are more accurate than LIPM, they need an iterative algorithm to define the COM trajectory due to the dependency of masses' motions on each other. 

Shimmyo et al.~\cite{shimmyo2013biped} extended the LIPM by considering three masses to decrease the modeling error and improve the performance. In their model, the masses were located on the torso, the right leg, and the left leg. They made two assumptions to design a preview controller which were considering the constant vertical position for the masses and also constant mass distribution. The performance of their approach has been validated through several experiments.

Faraji and Ijspeert~\cite{faraji20173lp} proposed a dynamics model which is composed of three linear pendulums to model the dynamics of the legs and the torso and they named it 3LP. They argued that, due to the linearity of this model, it is fast and computationally similar to LIPM such that it can be used in modern control architectures from the computational perspective. The performance of their model has been proven using a set of simulations. Simulation results showed that their framework was able to generate a robust walking with a wide range of speeds without requiring off-line optimization and tuning of parameters. 

Several extended versions of LIPM have been proposed to consider the momentum around the COM~\cite{pratt2006capture,kasaei2017reliable} but they do not take into account the dynamics of the legs specifically. In these models, the legs are considered to be mass-less and their motions do not have any effect in dynamics. In our previous work~\cite{kasaei2018optimal}, we extended the dynamics model presented in~\cite{pratt2006capture} by considering the mass of the stance leg, and we explained how this model can be represented by a first-order differential equation which can be used to design a controller to plan and track the walking reference trajectories. 

\subsection{Locomotion Framework}

Herdt et al.~\cite{herdt2010online} proposed an online walking generator with automatic footstep placement. They used LIPM to formulate the walking pattern generator as a Linear MPC. In their framework, instead of considering a set of predefined footsteps, they introduce new control variables to generate footsteps automatically. The performance of their approach has been validated through a set of simulations. The simulation results showed that the framework is able to track a given reference speed of COM that can be modified any time.

Brasseur et al.~\cite{brasseur2015robust} designed a robust linear MPC to generate online 3D locomotion for humanoid robots in a single computation with guaranteed kinematic and dynamic feasibility. They used Newton and Euler equations of motion to model the dynamics of a humanoid robot. They considered some assumptions to reduce the complexity of the model which are almost the same as LIPM's assumptions, but unlike LIPM, they take into account the vertical motion of the COM to generate more efficient locomotion in terms of energy and speed. The performance of their approach has been tested in two simulation scenarios including walking on a flat train and climbing stairs using a simulated HRP-2 robot. The simulation results validated the performance of their approach. 

Jianwen Luo et al.~\cite{luo2019robust} proposed a hierarchical control structure to generate robust biped locomotion. Their structure is composed of three independent control levels. At the highest level, they used $A^*$ to plan a path to the destination, and then they designed a MPC based on LIPM to generate the COM trajectory. At the middle level, they designed a hybrid controller which is composed of two controllers to control the oscillation and to eliminate the chattering problem. At the lowest level, they used a whole body operational space~(WBOS) controller to compute the joint torques analytically. The effectiveness of their approach has been validated through simulation.

% Chang et al.~\cite{chang2015humanoid} proposed a push recovery strategy based on centroidal-moment-pivot~(CMP) criterion and angular momentum regulation. In their approach, they formulated COG angular momentum regulator as a mass-damper-spring system and by defining a threshold, they activate their step controller to modify the swing leg trajectory. The performance of their approach has been validated through simulations and experiments and the results validated the effectiveness of their strategy. 

% Chenglong Fu~\cite{6942904} proposed a perturbation recovery strategy based on updating the footstep. This strategy is composed of two stages, a desired footstep calculator and a swing leg controller. In the desired footstep calculator, the acceleration and the jerk of the robot's hip is constantly monitored to detect the perturbation and to quickly modify the desired step location along with a capturability inspection. The swing leg controller is responsible for generating the joint trajectories and torques according to the updated footstep. The performance and effectiveness of his method have been validated through simulation using a $45.3Kg$ simulated robot. The simulation results showed that the robot was able to recover from an impulse of $20Ns$.

Most of the aforementioned works use LIPM as their dynamics model and do not take into account the dynamics of the swing leg and the torso. In this paper, we propose a modular framework to generate a robust biped locomotion which takes into account the dynamics of torso and swing leg. Particularly, we use the three-mass model as the core of this framework and based on that, we design a MPC controller to formulate walking as a set of quadratic functions and a set of time-varying constraints based on states, inputs, and outputs of the system. This controller is not only able to track the reference trajectories optimally even in presence of measurement noise but also it is robust against uncertainties such as external disturbances. Furthermore, we will performed a set of simulations to explore the impacts of the proposed controller and to validate the performance of the framework.

\begin{figure}[t]
	\begin{centering}
		\begin{tabular}	{c}	
			\includegraphics[width=0.95\linewidth,trim= 0cm 0cm 0cm 0cm,clip] {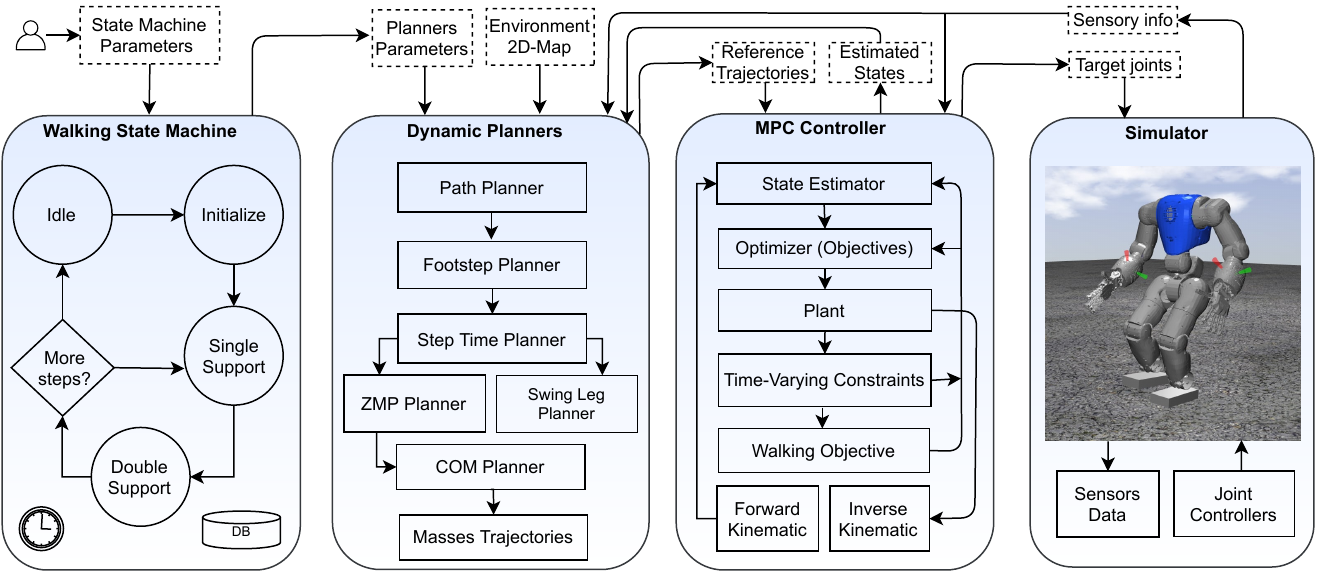}		
		\end{tabular}	
	\end{centering}			
	\caption{Overall architecture of the proposed framework. This framework is composed of four main modules which are connected together to generate robust locomotion. The highlighted boxes represent the functional blocks and the exchanged information among the modules are represented by the white boxes.  }
	%	\vspace{-2mm}
	\label{fig:overall}
\end{figure}

\section{Architecture}
\label{sec:arct}
This section is focused on presenting the overall architecture of the proposed framework which is depicted in Fig.~\ref{fig:overall}. This framework is composed of four main modules which are coupled together to generate robust biped locomotion. The first module is \texttt{Walking State Machine} which is responsible for controlling the overall process of the walking. According to the periodic nature of the walking, this state machine is composed of four main states such that state transitions are triggered based on an associated timer and also state conditions. In the \textit{Idle state}, the robot stands and is waiting for a start walking command. Once a walking command is received, the state transits to the \textit{Initialize state} and the robot shifts its COM to the first support foot to be ready for taking the first step. In the \textit{Single Support State} and the \textit{Double Support State}, the planners parameters (e.g., robot's target position and orientation, max speed, etc. ) will be updated. These parameters along with a 2D map of the environment and the sensory information provided by the \texttt{Simulator} as well as the estimated states  provided by the \texttt{MPC Controller} will be fed into the \texttt{Dynamic Planner} to generate a set of reference trajectories. \texttt{Dynamic Planner} starts by generating a collision-free path which will be used to plan a set of reference trajectories including footsteps, ZMP, swing, COM, and masses trajectories. Finally, these references will be fed into the \texttt{MPC Controller}. These trajectories along with the sensory information (e.g., joint encoders, IMU, robot's position and orientation, etc.) provided by the \texttt{Simulator} will be used to estimate the robot's states. The estimated states and a set of objectives are used by the optimizer to specify the control inputs subject to a set of time-varying constraints for tracking the generated reference trajectories. The corresponding joint motions will be generated using the \textit{Inverse Kinematics} which takes into account kinematic feasibility constraints. The target joint setpoints will be fed into the \texttt{Simulator} which is responsible for simulating the interaction of the robot with the environment and producing sensory information as well as the global position and orientation of the robot in the simulated environment to be used by \texttt{Dynamic Planner}. In the following sections, the details of each module will be explained separately, and then a set of simulation scenarios will be designed and conducted to valid the performance of the framework.

\section{Dynamic Planners}
\label{sec:ReferenceTrajectories}
Dynamic Planners module is responsible for planning of the reference trajectories according to the input parameters as well as the environment's structure. To reduce the complexity of the planning, the planning process is divided into a set of sub-planners which are solved separately and connected in a hierarchical manner (see Fig.~\ref{fig:overall}). In the rest of this section, we will explain how the reference trajectories will be generated.

\subsection{A* Footstep Planning}
Footstep planning is generally based on a graph search algorithm with a rich history~\cite{hornung2012adaptive,griffin2019footstep}. In our target framework, the footstep planning is composed of two stages: \textit{(i)} generating a collision-free 2D body path; \textit{(ii)} planning the footsteps based on the generated path. In the first stage, the environment is modeled as a 2D grid map consisting of cells that are marked as free or occupied. In this work, the size of the cell is considered to be $0.1m^2$, the height of obstacles is not considered, and the robot can not step over them. In order to avoid collision with the obstacles, the size of the obstacles are considered larger than the real size (scale = 1.1). In this stage, after modeling the environment, the A* search method is applied to find an optimum path over the free cells from the current position of the robot to the goal. Euclidean distance to the goal is used to guide the search. In the second stage, the footsteps should be generated according to the generated path. To do so, a state variable is defined to describe the current state of the robot's feet:
\begin{equation}
s = (x_l, y_l, \theta_l,\phi_l, x_r, y_r, \theta_r, \phi_r)
\end{equation}
\noindent
\begin{figure}[t]
	\centering
	\begin{tabular}	{c}	
		\includegraphics[scale=0.55,trim= 0cm 0cm 0cm 0cm,clip] {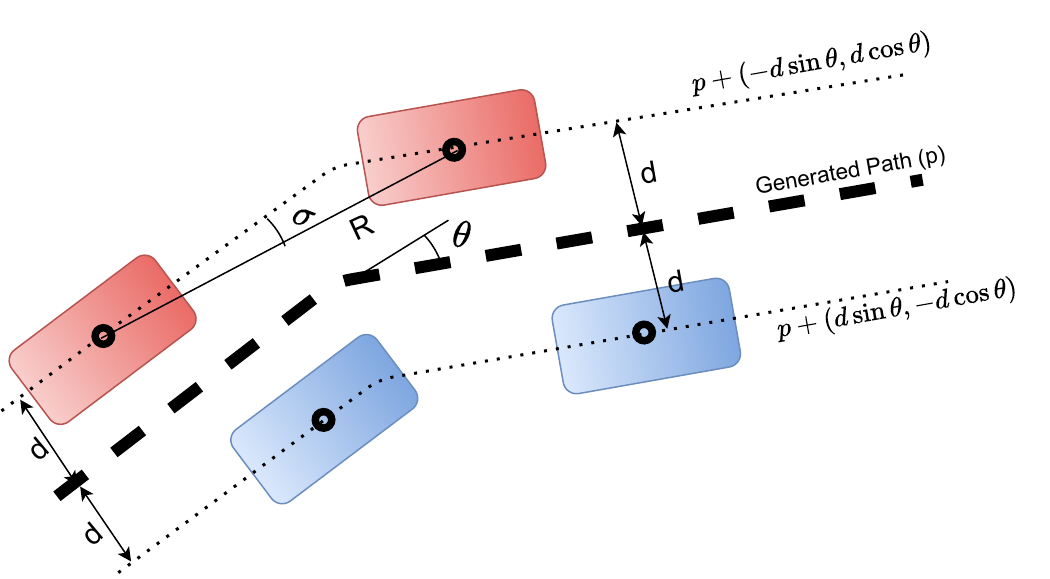}	
	\end{tabular}	
	\caption{Example of planning footsteps. After generating a collision-free path (thick dashed black line), two lines parallel to the path will be generated (dotted black lines) which are $d$~m away from the generated path where $2d$ is the distance between feet. Then, the left and right footsteps (red and blue rectangles) should be selected from these path so that $R$ m away from its previous one. Afterwards, $\sigma$ will be determined based on the generated footsteps using inverse tangent function.}
	%	\vspace{-2mm}
	\label{fig:footstep}
\end{figure}
where $x_l, y_l, \theta_l, x_r, y_r, \theta_r$ represent the current position and orientation of the left and the right feet, respectively. $\phi_l, \phi_r$ represent the state of each foot which is $1$ if the foot is the swing foot, and $-1$ otherwise. According to the nature of walking which is generated by moving the right and left legs, alternating, a step action is parameterized by a distance and an angle from the swing foot position at the beginning of step $a = (R, \sigma)$. Based on the current state and the generated path, action should be taken. In this paper, we consider a constant step size ($R = 0.1$m) which is a kind of medium step size that has been determined based on the robot capability and $\sigma$ is determined according to the generated path and the current state of the feet. Fig~\ref{fig:footstep} shows an example of planning footsteps after generating the collision-free path.

It is worth mentioning that in some scenarios like tele-operation tasks, an operator wants to determine the actions without any autonomous path planning. In such scenario, the step positions will be planned just according to the input command. It should be mentioned that the action is always passed through a first-order lag filter to ensure a smooth update. By applying the selected action, the state transits to a new state, $s' = t(s , a)$. It should be noted that after each transition, the current footstep will be saved ($f_i\quad i \in \mathbb{N} $), also  $\phi_l$ and $\phi_r$ will be toggled. After generating the footsteps, the step time should be planned based on the robot's target speed.

\subsection{ZMP, Hip and Swing Reference Trajectories}
Studies on human locomotion showed that while a human is walking, ZMP moves from heel to the toe during the single support phase and it moves towards the COM during the double support phase~\cite{dasgupta1999making,erbatur2002study}. In this paper, we do not consider the ZMP movement during the single support, but instead, we keep it in the middle of the support foot to prevent it from approaching to the support polygon edge and to avoid the loss of equilibrium of the robot in some situation like unexpected external disturbances. Thus, the ZMP reference trajectory planning can be formulated based on the generated footsteps as follows:
\begin{equation}
p_{st}= 
\begin{cases}
f_{i}  \qquad\qquad\qquad\qquad\qquad 0 \leq t < T_{ss} \\
f_{i}+ \frac{SL \times (t-T_{ss})}{T_{ds}} \qquad\qquad T_{ss} \leq t < T_{ss}+T_{ds} 
\end{cases} ,
\label{eq:zmpEquation}
\end{equation}
\noindent
where $p_{st} = [p_{st}^x \quad p_{st}^y]^\top$ is the generated ZMP, $t, T_{ss}, T_{ds}$ represent the time, duration of single and double support phases, respectively. $SL = [SL^x \quad SL^y]$ is a vector that represents the step length and step width which are determined based on $(R, \sigma)$, $f_{i} = [f_{i}^x \quad f_{i}^y]$ represents the planned foot steps on a 2D surface. It should be noted that $t$ will be reset at the end of each step~\mbox{$(t \geq T_{ss}+T_{ds}$)}.

After generating the ZMP reference trajectory, the hip trajectory will be planed according to the generated ZMP. To do that, we assume that the COM of the robot is located at the middle of the hip, and, based on this assumption, the overall dynamics of the robot is firstly restricted into COM and then the reference trajectory for the hip will be generated using the analytical solution of the LIPM as follows:

\begin{equation}
\label{eq:com_traj_x0xf}
\resizebox{0.75\linewidth}{!}{$ p_h(t) = p_{st} + \frac{ (p_{st}-p_{h_f}) \sinh\bigl((t - t_0)\omega\bigl)+ (p_{h_0} - p_{st}) \sinh\bigl((t - t_f)\omega\bigl)}{\sinh((t_0 - t_f)\omega)} $},
\end{equation}
\noindent
where $t_0$ and $t_f$ represent the beginning and the ending times of a step, $\omega = \sqrt{\frac{g}{l_h}} $ is the natural frequency of LIPM and $p_{h_0}$, $p_{h_f}$ are the corresponding positions of COM at these times, respectively.

After generating the ZMP and the hip trajectory, the swing trajectory should be planned. To have a smooth trajectory during lifting and landing of the swing leg, a Bézier curve is used to generate this trajectory according to the generated footsteps and a predefined swing height.
\begin{figure}[!t]
	\centering
	\includegraphics[width = 0.9\columnwidth, trim= 0.0cm 0.25cm 0cm 0cm,clip]{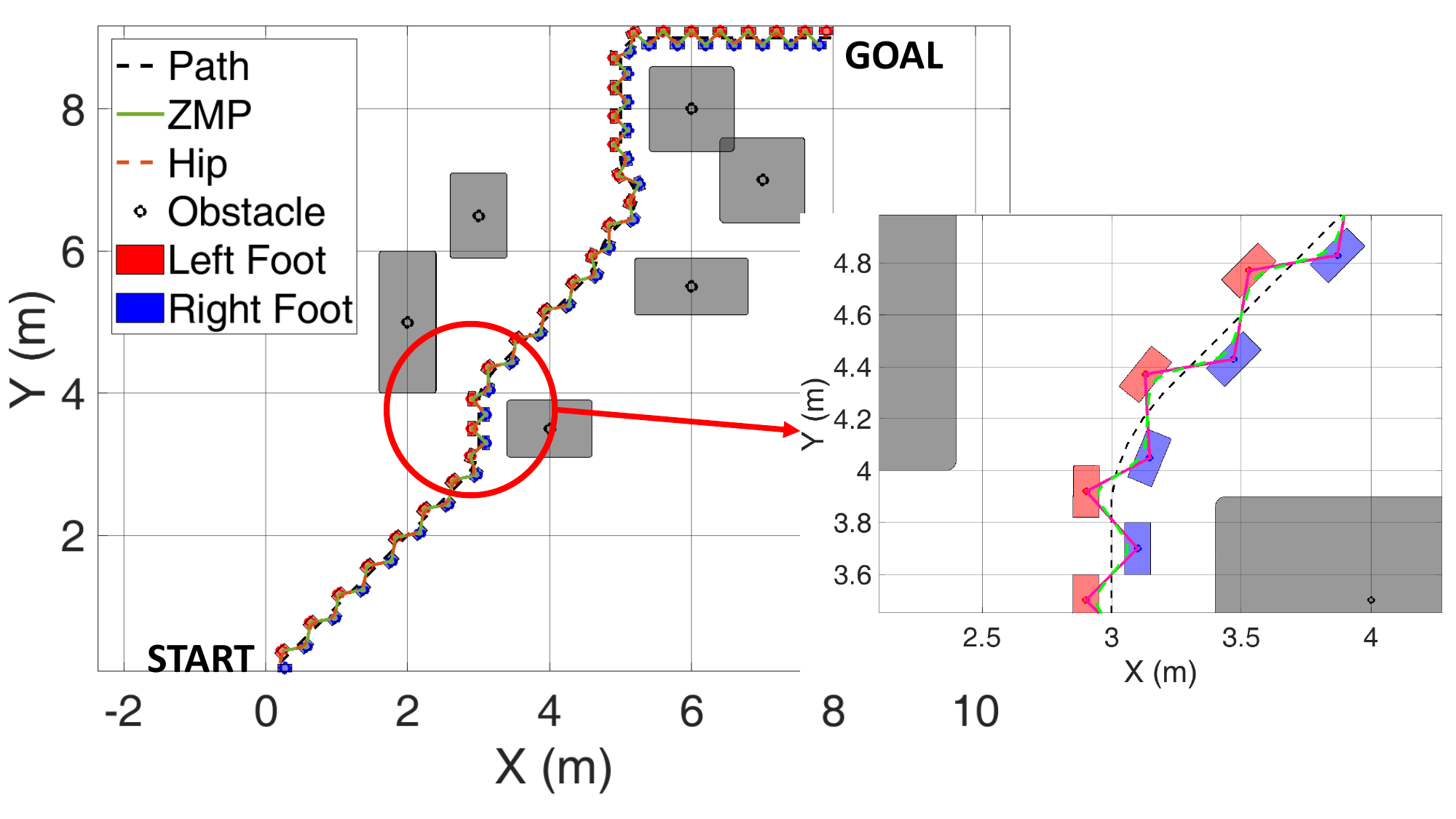} 
	\caption{ An example walking references trajectories: the gray rectangles represent the occupied cells that robot can not step; the black-dashed line represents the output of the A* path planner; red and blue rectangles represent the footsteps which are generated based on the output of the A*; the magenta line is the ZMP which is generated based on the outputs of the footstep planner; the lime-dashed line is the hip reference trajectory.}
	\vspace{-0mm}
	\label{fig:PathPlanning}
\end{figure}

\subsection{Masses Reference Trajectories}
The trajectories of the masses can be easily generated based on the geometric relations between the generated hip, ZMP, and swing leg trajectories. According to Fig.\ref{fig:dynamicsmodel}(b), the mass of the stance leg is located in the middle of the line between the ZMP and the hip. Similarly, the mass of the swing leg is located in the middle of the line between the swing foot and the hip. To show the performance of the planners presented in this section, an example path planning scenario has been set up, which is depicted in Fig.~\ref{fig:PathPlanning}. In this scenario, the robot stands at the \textit{START} point and wants to reach the \textit{GOAL} point. The planning process is started by generating an optimum obstacle-free path (black-dashed line). As it is shown in~Fig.~\ref{fig:PathPlanning}, the generated path is far from the obstacles enough and, based on this path, the footsteps, corresponding ZMP and hip trajectories have been generated successfully.

\section{Dynamics Model}
\label{sec:DynamicsModel}
This section will be started by a brief review of the Zero Momentum Point~(ZMP) concept which is one of the well-known criteria in developing stable dynamic walking. Afterwards, the ZMP concept will be used to define the overall dynamics of a biped robot as a state-space system.

\subsection{Zero Momentum Point and Gait Stability}
Several criteria have been introduced to analyze the stability of a biped robot. ZMP is a well-known criterion to develop dynamically stable locomotion and its popularity comes from its performance and ease of use. In fact, ZMP is the point on the ground plane where the ground reaction force~(GRF) acts to cancel the inertia and gravity~\cite{vukobratovic1970stability}. For a dynamics model which is composed of $n$ parts, ZMP can be calculated using the following equation:

\begin{equation}
p = \frac{\sum_{k=1}^{n} m_k c_k (\ddot{z}_{k} + g) - \sum_{k=1}^{n} m_k z_k \ddot{c}_k  }{\sum_{k=1}^{n} m_k (\ddot{z}_{k} + g)} \quad , 
\label{eq:zmp}
\end{equation}
\noindent
where $g$ represents the acceleration of gravity, $p = [p^x \quad p^y]^\top$ represents the position of ZMP, $m_k$ is the mass of each part, $c_k= [c_{k}^x \quad c_{k}^y]^\top$ and $\ddot{c}_k$ denote the ground projection of the position and acceleration of each mass, $z_k, \ddot{z}_k$ are vertical position and vertical acceleration of each mass, respectively. 

\subsection{Three-mass Dynamics Model}
The three-mass model abstracts the dynamics model of a biped robot by considering three masses. As shown in Fig.~\ref{fig:dynamicsmodel}(b), the masses are placed at the legs and the torso of the robot. To simplify and linearize the model, each mass is restricted to move along a horizontal plane. According to this assumption, the ZMP equations are independent and equivalent in the frontal and sagittal planes. Thus, in the remainder of this paper, just the equations in the sagittal plane will be considered. Based on~(\ref{eq:zmp}), a state-space system can be defined to analyze the behavior of the system:
\begin{equation}
\begin{gathered} 
\resizebox{0.6\linewidth}{!}{$
	\frac{d}{dt} \underbrace{\begin{bmatrix} c_{1}^{x} \\ \dot{c}_{1}^{x} \\ \ddot{c}_{1}^{x} \\ c_{2}^{x} \\ \dot{c}_{2}^{x} \\ \ddot{c}_{2}^{x} \\c_{3}^{x} \\ \dot{c}_{3}^{x} \\ \ddot{c}_{3}^{x} \end{bmatrix}}_{X}
	= 
	\underbrace{\begin{bmatrix} % A 
		0 & 1 & 0 & 0 & 0 & 0 & 0 & 0 & 0  \\
		0 & 0 & 1 & 0 & 0 & 0 & 0 & 0 & 0  \\
		0 & 0 & 0 & 0 & 0 & 0 & 0 & 0 & 0  \\
		0 & 0 & 0 & 0 & 1 & 0 & 0 & 0 & 0  \\
		0 & 0 & 0 & 0 & 0 & 1 & 0 & 0 & 0  \\
		0 & 0 & 0 & 0 & 0 & 0 & 0 & 0 & 0  \\
		0 & 0 & 0 & 0 & 0 & 0 & 0 & 1 & 0  \\
		0 & 0 & 0 & 0 & 0 & 0 & 0 & 0 & 1  \\
		0 & 0 & 0 & 0 & 0 & 0 & 0 & 0 & 0  
		\end{bmatrix}}_{A}	
	\underbrace{\begin{bmatrix} c_{1}^{x} \\ \dot{c}_{1}^{x} \\ \ddot{c}_{1}^{x} \\ c_{2}^{x} \\ \dot{c}_{2}^{x} \\ \ddot{c}_{2}^{x} \\c_{3}^{x} \\ \dot{c}_{3}^{x} \\ \ddot{c}_{3}^{x} \end{bmatrix}}_{X}
	+
	\underbrace{\begin{bmatrix} %B
		0 & 0 & 0 \\
		0 & 0 & 0 \\
		1 & 0 & 0 \\
		0 & 0 & 0 \\
		0 & 0 & 0 \\
		0 & 1 & 0 \\
		0 & 0 & 0 \\
		0 & 0 & 0 \\
		0 & 0 & 1 
		\end{bmatrix}}_{B} \underbrace{\begin{bmatrix} \dddot{c}_1^{x} \\ \dddot{c}_2^{x} \\ \dddot{c}_3^{x} \end{bmatrix}}_{U}$}
\\
\resizebox{0.6\linewidth}{!}{$
	y = \underbrace{\begin{bmatrix} %C
		1 &0 &0 &0 &0 &0 &0& 0& 0& \\
		%0 &0 &0 &1 &0 &0 &0& 0& 0& \\
		0 &0 &0 &0 &0 &0 &1& 0& 0& \\
		\frac{m_1}{M} &0 &\frac{-m_1 z_1}{M g} & \frac{m_2}{M} &0 &\frac{-m_2 z_2}{M g} & \frac{m_3}{M} &0 &\frac{-m_3  z_3}{M g}
		\end{bmatrix}}_{C}
	\underbrace{\begin{bmatrix} c_{1}^{x} \\ \dot{c}_{1}^{x} \\ \ddot{c}_{1}^{x} \\ c_{2}^{x} \\ \dot{c}_{2}^{x} \\ \ddot{c}_{2}^{x} \\c_{3}^{x} \\ \dot{c}_{3}^{x} \\ \ddot{c}_{3}^{x} \end{bmatrix}}_{X} , $}
\label{eq:statespace_3lipm}
\end{gathered} 
\end{equation}
\noindent
where $\dddot{c}_1^{x},\dddot{c}_2^{x},\dddot{c}_3^{x}$ are the  manipulated variables in jerk dimension, $M$ is weight of the robot, $m_1, m_2, m_3$ represent the mass of stance leg, torso and swing leg, respectively. Based on the output equation~(y), the positions of stance leg, swing leg, and also ZMP are measured at each control cycle. In the next section, the defined system will be discretized to discrete-time implementation and we will explain what type of walking objectives and constraints should be considered to generate stable dynamic walking.

\section{Online Walking Controller Based on MPC}
\label{sec:MPC}
In this section, the problem of the online walking controller is formulated as a linear MPC which is not only robust against uncertainties but also able to consider some constraints in the states, inputs and outputs. To do that, firstly, the presented continuous system~(\ref{eq:statespace_3lipm}) should be discretized for implementation in discrete time. Afterward, the walking objective will be formulated as a set of quadratic functions and finally, walking constraints will be formulated as linear functions of the states, inputs, and outputs. 

\subsection{Discrete Dynamics Model}
To discretize the system, we assume that $\ddot{c}_1, \ddot{c}_2, \ddot{c}_3$ are linear and, based on this assumption, $\dddot{c}_1, \dddot{c}_2, \dddot{c}_3$ are constant within a control cycle. Thus, the discretized system can be represented as follows:
\begin{equation}
\begin{gathered} 
X(k+1)=A_d X(k)+B_d u(k)\quad \\
y(k) = C_d X(k)
\end{gathered} 
%\end{aligned}
\label{eq:statespace_lipm_disc} 
\end{equation}
\noindent
where $k$ represents the current sampling instance, $A_d, B_d, C_d$ are the discretized version of the $A, B, C$ matrices in~(\ref{eq:statespace_3lipm}), respectively. Based on this system, the state vector $X(k)$ can be estimated at each control cycle. Thus, according to the estimated states, some objectives, and constraints, the problem of determining the control inputs can be formulated as an optimization problem~(a quadratic program~(QP)) at each control cycle. The optimization solution specifies the control inputs which are used until the next control cycle. This optimization just considers the current timeslot, to take into account the future timeslots, a finite time horizon is considered for the optimization process, but only the current timeslot will be applied and for each timeslot, this optimization will be repeated. Indeed, the optimization solution determines $N_c$ (control horizon) future moves ($\Delta U =[ \Delta u(k), \Delta u(k+1),...,  \Delta u(k+N_c-1)]^\top$) based on the future behavior of the system ($Y = [y(k+1|k),y(k+2|k),...,y(k+N_p|k)]^\top$) over a prediction horizon of $N_p$.

\subsection{Walking Objective}
Walking is a periodic locomotion which can be decomposed into two main phases: \textit{single support} and \textit{double support}. In the double support phase,  the robot shifts its COM to the stance foot and during the single support, its swing leg moves towards the next step position. In order to develop stable locomotion, the robot should be able to track a set of reference trajectories while keeping its stability. As mentioned before, a popular approach to guarantee the stability of the robot is keeping the ZMP within the support polygon. According to~(\ref{eq:statespace_3lipm}), the position of the stance leg, swing leg, and ZMP are measured at each control cycle. Thus, the following objectives should be considered to keep the outputs at or near the references:
\begin{equation}
\begin{aligned}
d_1 &= || p_z - r_z||^2 \\
d_2 &= || p_{st} - r_{st}||^2 \\
d_3 &= || p_{sw} - r_{sw}||^2 ,
\end{aligned}
\end{equation}
\noindent
where $p_z, r_z, p_{st}, r_{st}, p_{sw}, r_{sw}$ are measured and reference ZMP, and the measure and reference positions of the stance and swing legs, respectively. Moreover, to generate a smooth trajectories which are compatible with the robot structure, the control inputs are considered into the objectives for smoothing all motions:
\begin{equation}
\begin{aligned}
d_4 &= || \dddot{c}_1^x||^2 \\
d_5 &= || \dddot{c}_2^x||^2 \\
d_6 &= || \dddot{c}_3^x||^2 .
\end{aligned}  
\end{equation}

According to the explained objective terms, the following cost function is defined to find an optimal set of control inputs:
\begin{equation}
J(z_k) = \sum_{i=1}^{N_p} \sum_{j=1}^{6}\alpha_j d_j(z_k) 
\label{eq:costfunction}
\end{equation}
\noindent
where $k$ is the current control interval, $z_k^\top = \{\Delta u(k|k)^\top\quad \Delta u(k+1|k)^\top  ... \Delta u(k+N_p-1|k)^\top \}$ is the QP decision and  $\alpha_j$ represents a positive gain that is assigned to each objective.

\begin{figure}[!t]
	\centering
	\includegraphics[width = 0.95\columnwidth, trim= 0.0cm 2.5cm 0cm 0cm,clip]{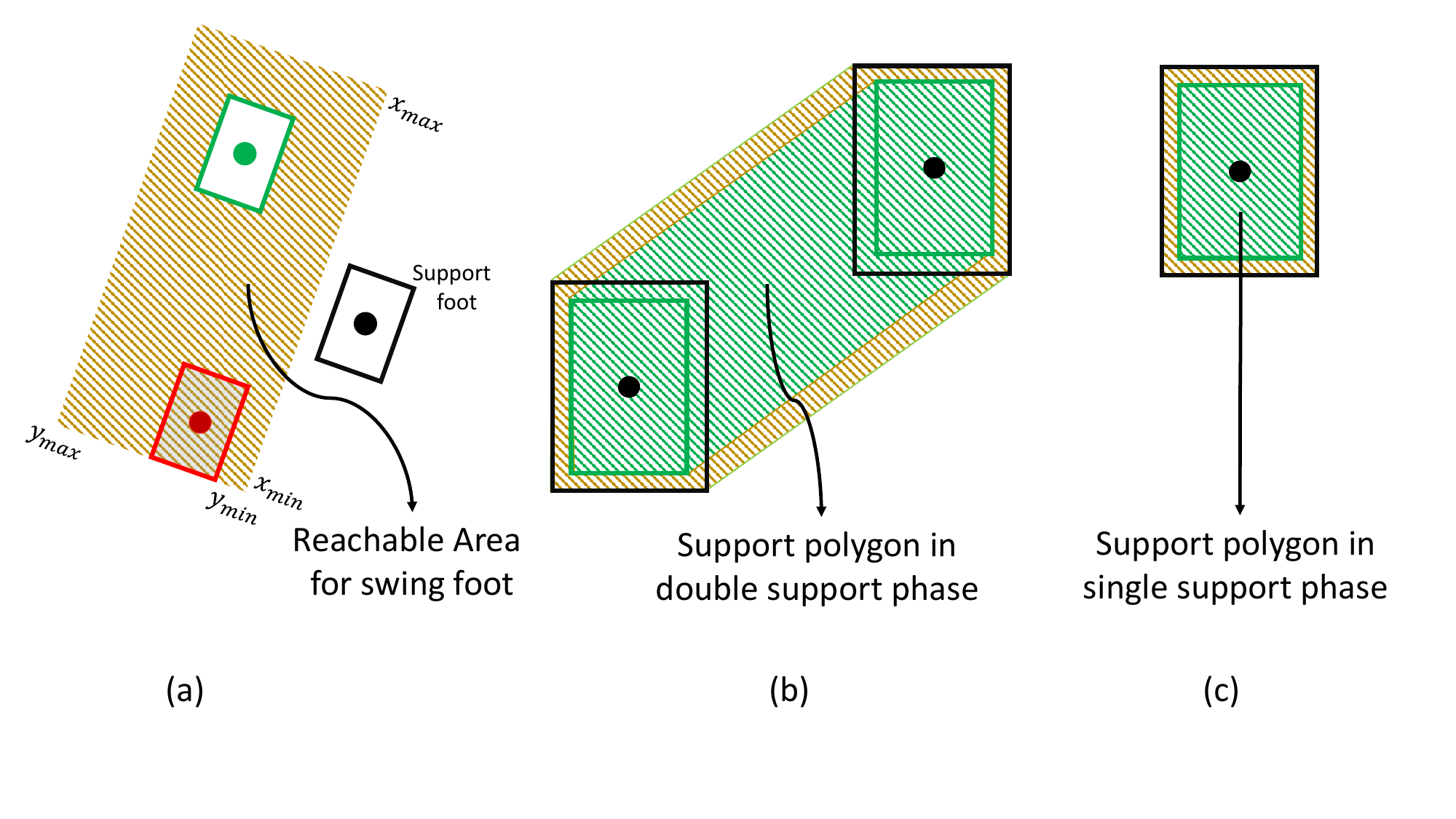} 
	\caption{ Graphical representations of the constraints: \textbf{(a)} kinematically reachable area for the swing leg: red rectangle represents the position of the swing leg at the beginning of step, green rectangle represents the landing location of the swing leg; \textbf{(b)} ZMP constraint during double support phase: green area is considered as the support polygon to avoid ZMP from being close to the borders; \textbf{(c)} ZMP constraint during single support phase.}
	\vspace{-0mm}
	\label{fig:ConstraintsVisualization}
\end{figure}
\subsection{Time-Varying Constraints}
To ensure the feasibility of the solution that is found by the MPC, a set of constraints should be considered to avoid generating infeasible solutions. For instance, the solution should be kinematically reachable by the robot and the ZMP should be kept inside the support polygon. Generally, a set of mixed input/output constraints can be specified in the following form:
\begin{equation}
Eu(k + j|k) + Fy(k + j|k) \le G + \epsilon
\label{eq:constraints}
\end{equation}
\noindent
where $j$ from 0 to $N_p$, $k$ is current control cycle, $E, F, G$ are time-variant matrices, where each row represents a linear constraint, $u(k + j|k)$ and $u(k + j|k)$ are vectors of manipulated variables and output variables (stance leg, swing leg and ZMP), respectively,  $\epsilon$ is used to specify a constraint to be soft or hard. Additionally, using this equation, the inputs and outputs can be bounded to specified limitations. It should be noted that the constraints in the sagittal plane are similar to those in the frontal plane.
% therefore, in the following of this section, just the constraints in the sagittal plane will be explained.
In our target framework, the constraints are time-varying and they will be determined at the beginning of each walking phase. Graphical representations of the constraints are depicted in Fig.~\ref{fig:ConstraintsVisualization}. Fig.~\ref{fig:ConstraintsVisualization}(a) shows that the landing position of the swing leg should be restricted in a kinematically reachable area. This area can be approximated by a rectangle which is defined as follows:

\begin{equation}
E =\begin{bmatrix} 0 &0 & 0\\0 & 0 & 0\end{bmatrix},  F=\begin{bmatrix} 0 & 1 & 0\\0 & -1 & 0\end{bmatrix},
G =\begin{bmatrix} x_{max}\\ x_{min} \end{bmatrix},  \epsilon = \begin{bmatrix}0 \\ 0\end{bmatrix},
\label{eq:SwingConstraints}
\end{equation}
\noindent
where $x_{min}, x_{max}$ are defined based on the current position of the support foot and the robot capability.  In addition to these constraints, to keep the ZMP within the support polygon, the following constraints are considered:
\begin{equation}
E =\begin{bmatrix} 0 &0 & 0\\0 & 0 & 0\end{bmatrix},  F=\begin{bmatrix} 0 & 0 & 1\\0 & 0 & -1\end{bmatrix},
G =\begin{bmatrix} z_{x_{max}}\\ z_{x_{min}}  \end{bmatrix},  \epsilon = \begin{bmatrix}0 \\ 0\end{bmatrix},
\label{eq:ZMPConstraints}
\end{equation}
\noindent
where $z_{x_{min}}, z_{x_{max}}$ are defined based on the walking phase and the size of the foot (please look at Fig.~\ref{fig:ConstraintsVisualization}. (b, c)). It should be mentioned that the foot size of the robot is considered a bit smaller (scale = 0.9) than the real foot size to prevent ZMP from being too close to the borders of the support polygon.

\section{Simulation}
\label{sec:simulation}
In this section, a set of simulation scenarios will be designed to validate the performance and examine the robustness of the proposed framework. We firstly performed three simulations using \mbox{MATLAB} to evaluate the performance and robustness of the framework. Afterward, we will deploy our framework on a simulated COMAN~(a passively compliant humanoid)~\cite{spyrakos2013push} humanoid robot to validate the performance of the proposed framework on a full-size humanoid robot.

\subsection{Simulation using MATLAB}
To perform the simulations using \mbox{MATLAB}, a humanoid robot has been simulated according to the dynamics model presented in Section~\ref{sec:DynamicsModel}. The most important parameters of the simulated robot and the controller are shown in Table~\ref{tb:Params}.
\begin{table}[h!]
	\centering
	\caption{Parameters used in the simulations.}
	\label{tb:Params}	
	\resizebox{\linewidth}{!}{
		\begin{tabular}{c | c |c| c | c |c | c|c |c |c|c|c} 		
			$m_1$&$m_2$ &  $m_3$ & $l_{st}$ & $l_t$&$l_{sw}$ & foot length & $T_s$&$N_p$&$N_c$&$\alpha_{1,2,3}$&$\alpha_{4,5,6}$ \\ %[0.5ex] 
			\hline			
			$15kg$ & $50kg$ & $15kg$ & $0.5m$ & $1.2m$ & $0.5m$ & $0.1m$ & $0.02s$ & $80$ & $20$ & $20$ & $0.1$ 
		\end{tabular}
	}
	\vspace{-4mm}
\end{table}

\subsubsection{Tracking Performance}
\begin{figure}[!t]
	\centering
	\begin{tabular}	{c c}	
		\includegraphics[width = 0.45\columnwidth, trim= 0.0cm 0.cm 0cm 0cm,clip]{figures/Figures/trackingall4.pdf} &
		\includegraphics[width = 0.45\columnwidth, trim= 0.0cm 0.cm 0cm 0cm,clip]{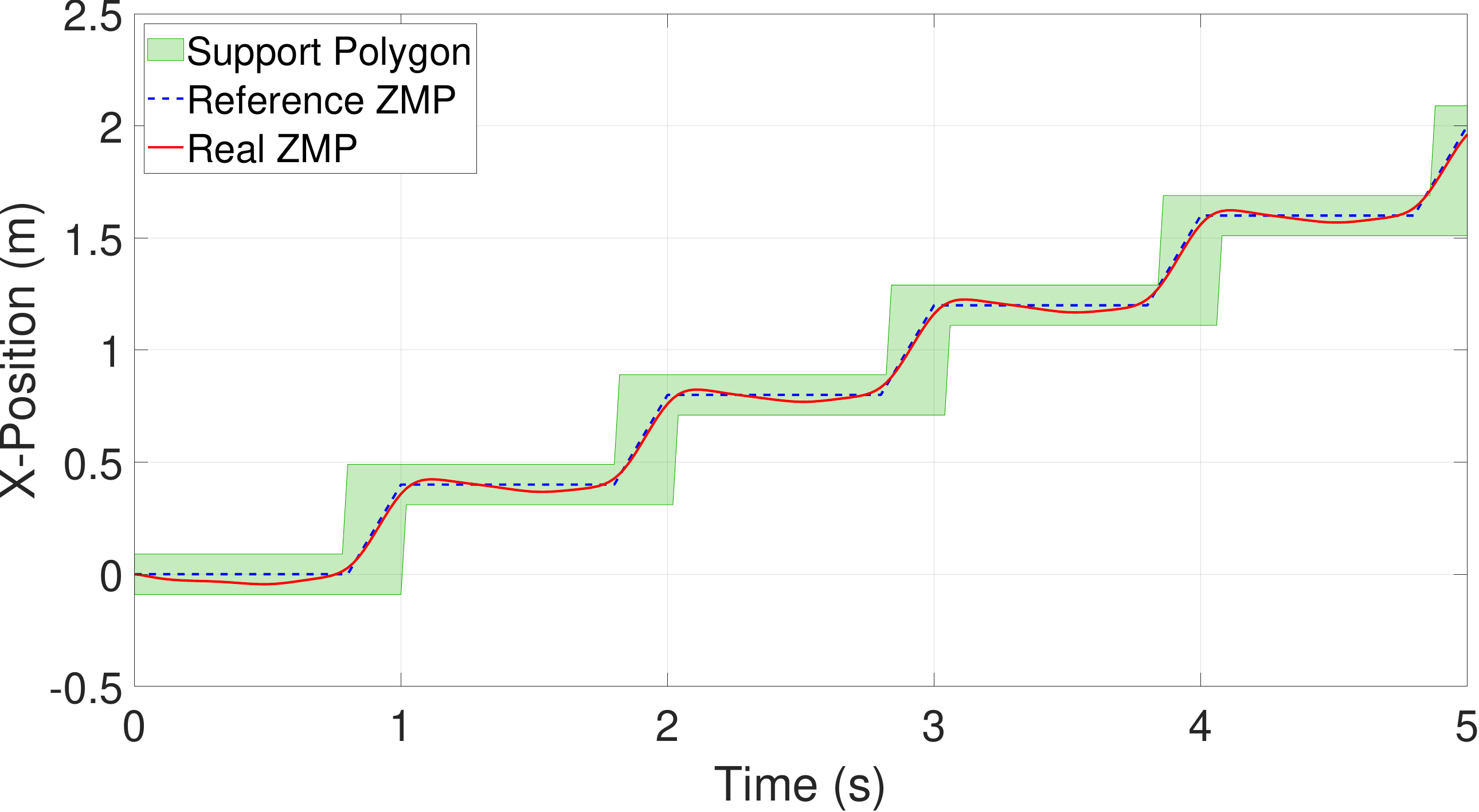}\\
		\textbf{(a)} & \textbf{(b)}
	\end{tabular}
	\vspace{-2mm}
	\caption{ The simulation results of analyzing tracking performance: \textbf{(a)} represents the positions of the masses while walking; \textbf{(b)} represents the reference ZMP and the real ZMP.}	
	\vspace{-0mm}
	\label{fig:tracking}
\end{figure}

To check the tracking performance of the controller, the simulated robot is commanded to perform a five-step forward walk (step size = $0.8m$ and step time = $1s$). In this simulation, the simulated robot is considered to be stopped and stands at the beginning of the simulation and the reference trajectories have been generated based on the presented methods in Section~\ref{sec:ReferenceTrajectories}. The exemplary planned trajectories at the end of Section~~\ref{sec:ReferenceTrajectories} are used as a the input references and the controller should track these trajectories while keeping the stability of the robot. The simulation results are depicted in Fig.~\ref{fig:tracking}. The results show that the controller is able to track the references and the ZMP is always inside the support polygon during walking. Another interesting point in the results is the actual trajectory of the torso (see $Real_{p_t}$ in Fig.~\ref{fig:tracking}). We did not determine any references for the torso explicitly, but it moves as we expected. As it is shown in Fig.~\ref{fig:tracking}\textbf{(a)}, it is almost near the support foot during the single support phase, then by starting the double support phase, it moves towards the next support foot.

\subsubsection{Robustness w.r.t. Measurement Noise}
\begin{figure}[!t]
	\centering
	\begin{tabular}	{c c}	
		\includegraphics[width = 0.45\columnwidth, trim= 0.0cm 0.cm 0cm 0cm,clip]{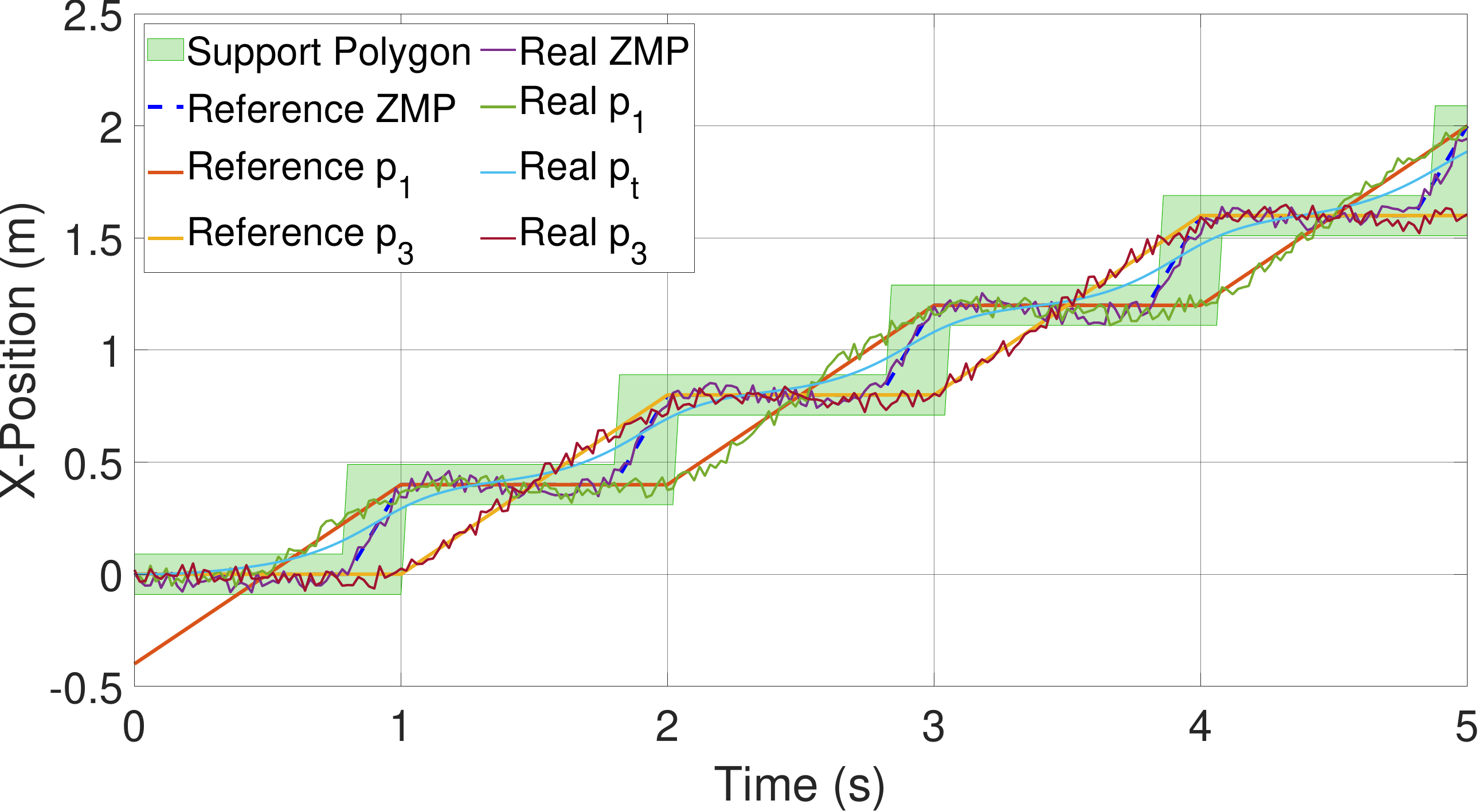} &
		\includegraphics[width = 0.45\columnwidth, trim= 0.0cm 0.cm 0cm 0cm,clip]{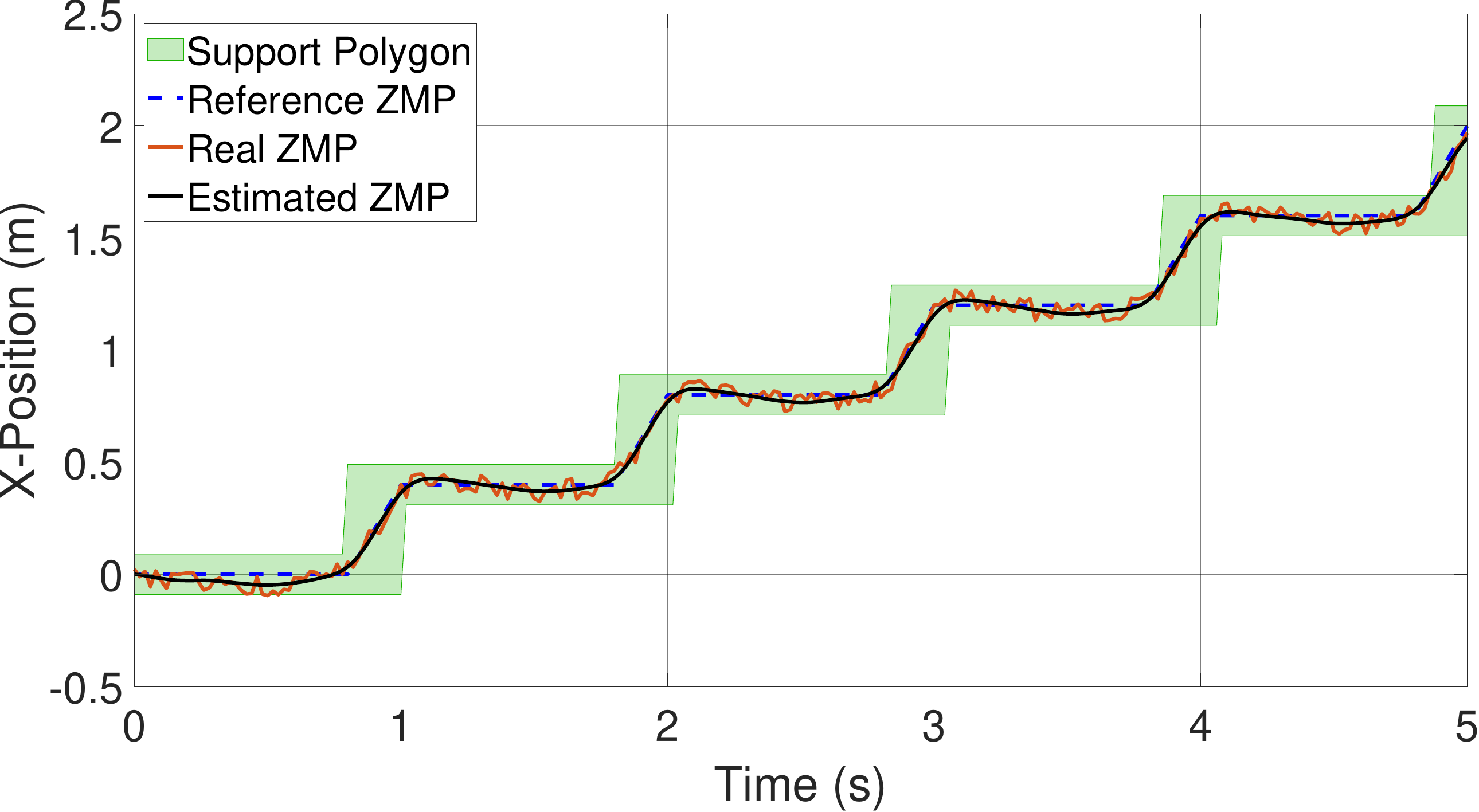} \\
		\textbf{(a)} & \textbf{(b)}
	\end{tabular}
	\vspace{-0mm}
	\caption{ The simulation results of examining the robustness w.r.t. measurement noise: \textbf{(a)} represents the real and estimated position of the masses; \textbf{(b)} represents the real and estimated ZMP.}	
	\vspace{-0mm}
	\label{fig:noise}
\end{figure}

In the real world, measurements are always affected by noise, therefore they are never perfect. Noise can arise because of many reasons like the simplification in the modeling, discretizing and some mechanical uncertainties~( e.g.,~backlash of gears), etc. A robust controller should be able to estimate the correct state using noisy measurements and minimize the effect of noise. To examine the robustness of the proposed controller regarding measurement noise, the measurements are modeled as a stochastic process by adding Gaussian noise ($-0.05m\le v_i\le0.05m\quad~i=1,2,3$) to the system output and the previous scenario has been repeated. The simulation results are shown in Fig.~\ref{fig:noise}. As can be seen, the controller is robust against the measurement noise and it can track the references even in the presence of noise.
\begin{figure}[!t]
	\centering
	\begin{tabular}	{c c}	
		F = 100N & F = -100N\\
		\includegraphics[width = 0.45\columnwidth, trim= 0.0cm 0.cm 0cm 0cm,clip]{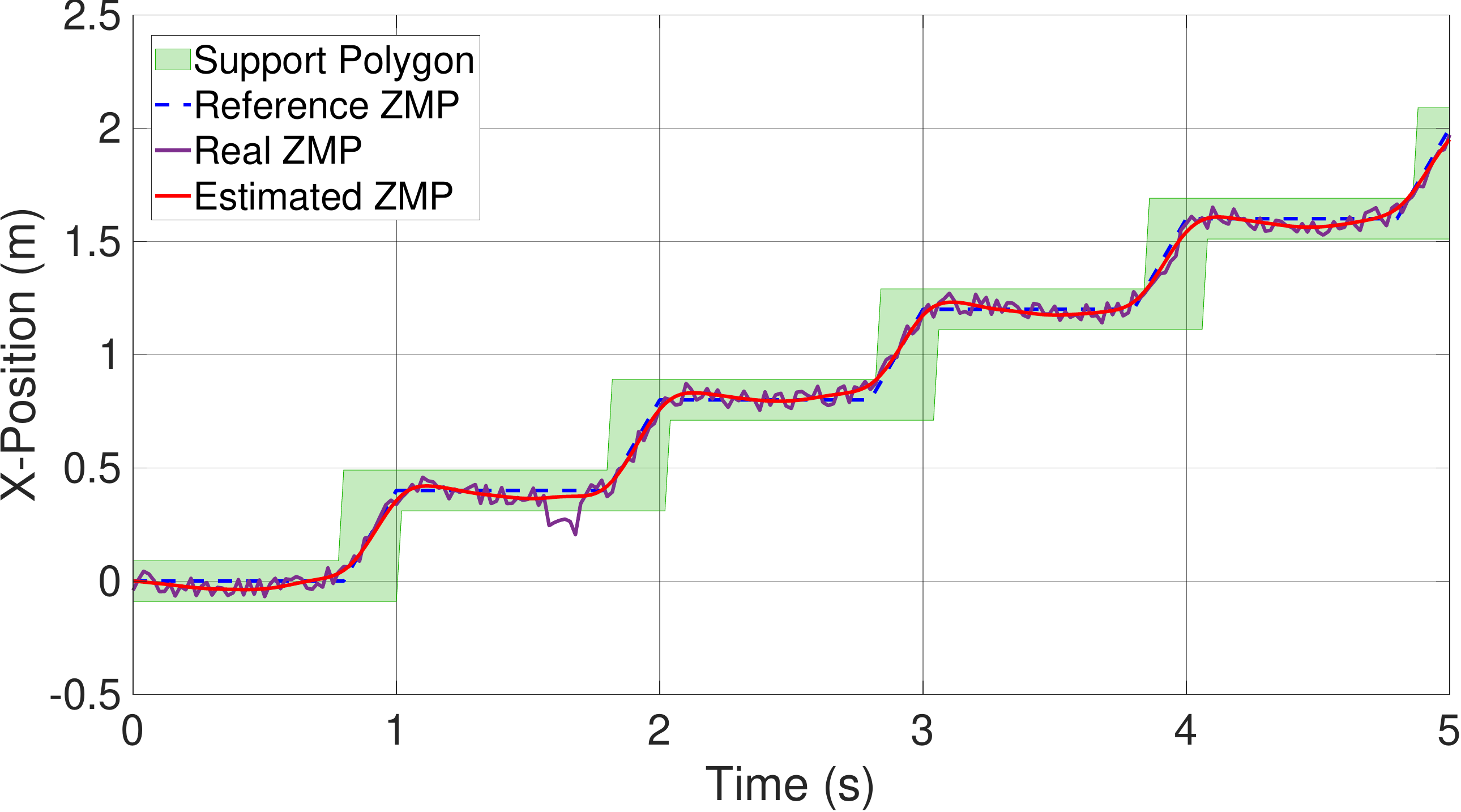} &
		\includegraphics[width = 0.45\columnwidth, trim= 0.0cm 0.cm 0cm 0cm,clip]{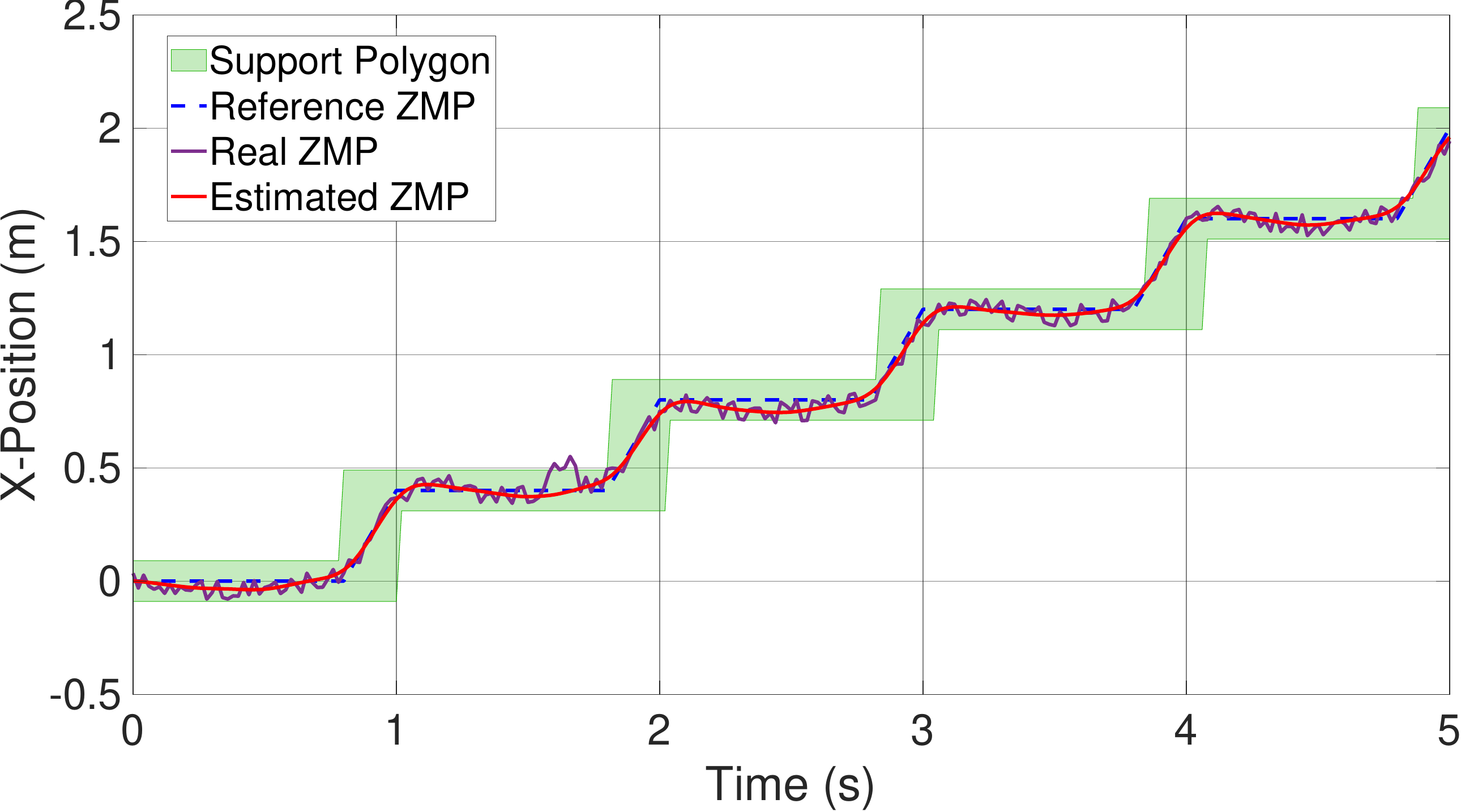}\\
		F = 200N & F = -200N\\
		\includegraphics[width = 0.45\columnwidth, trim= 0.0cm 0.cm 0cm 0cm,clip]{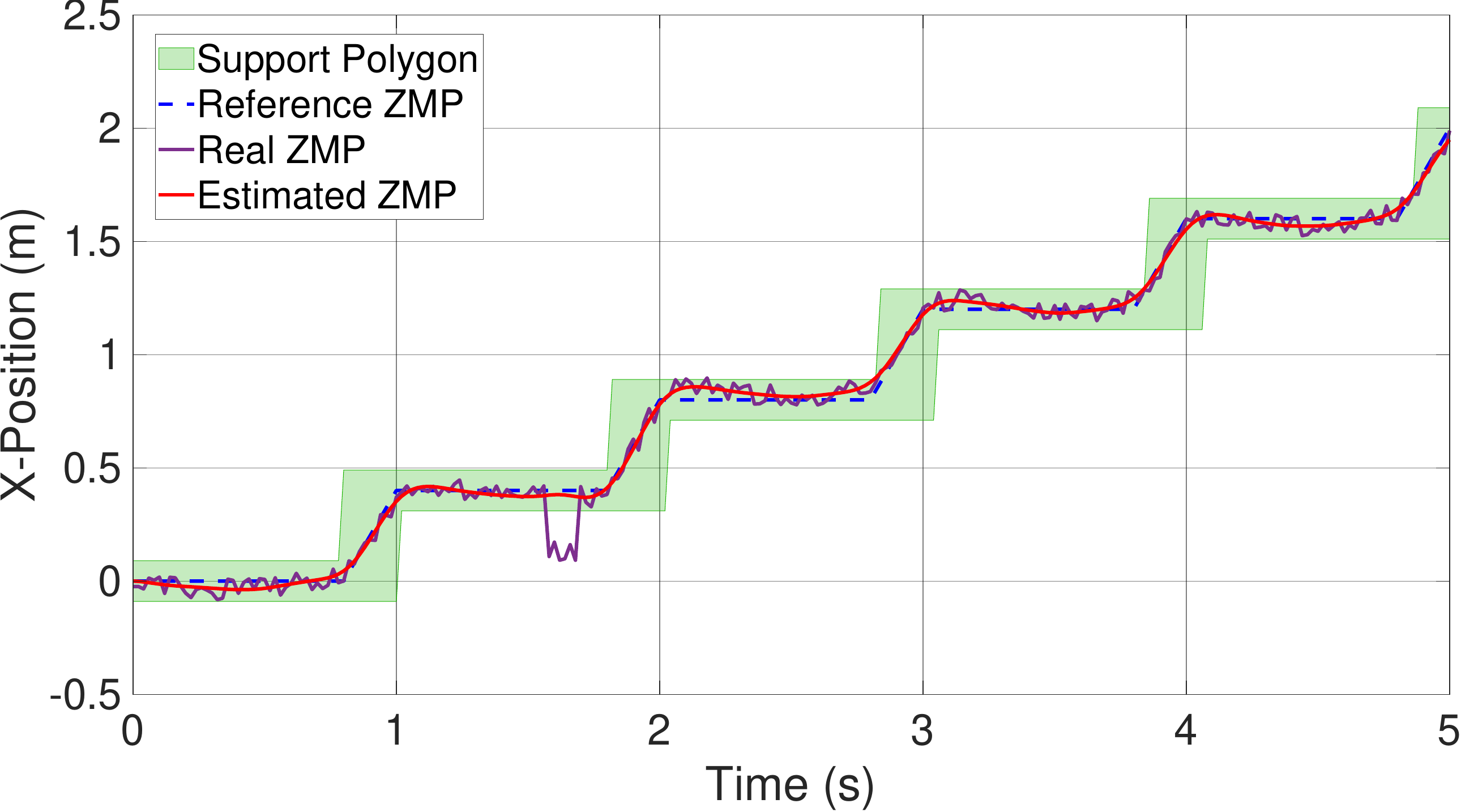}&
		\includegraphics[width = 0.45\columnwidth, trim= 0.0cm 0.cm 0cm 0cm,clip]{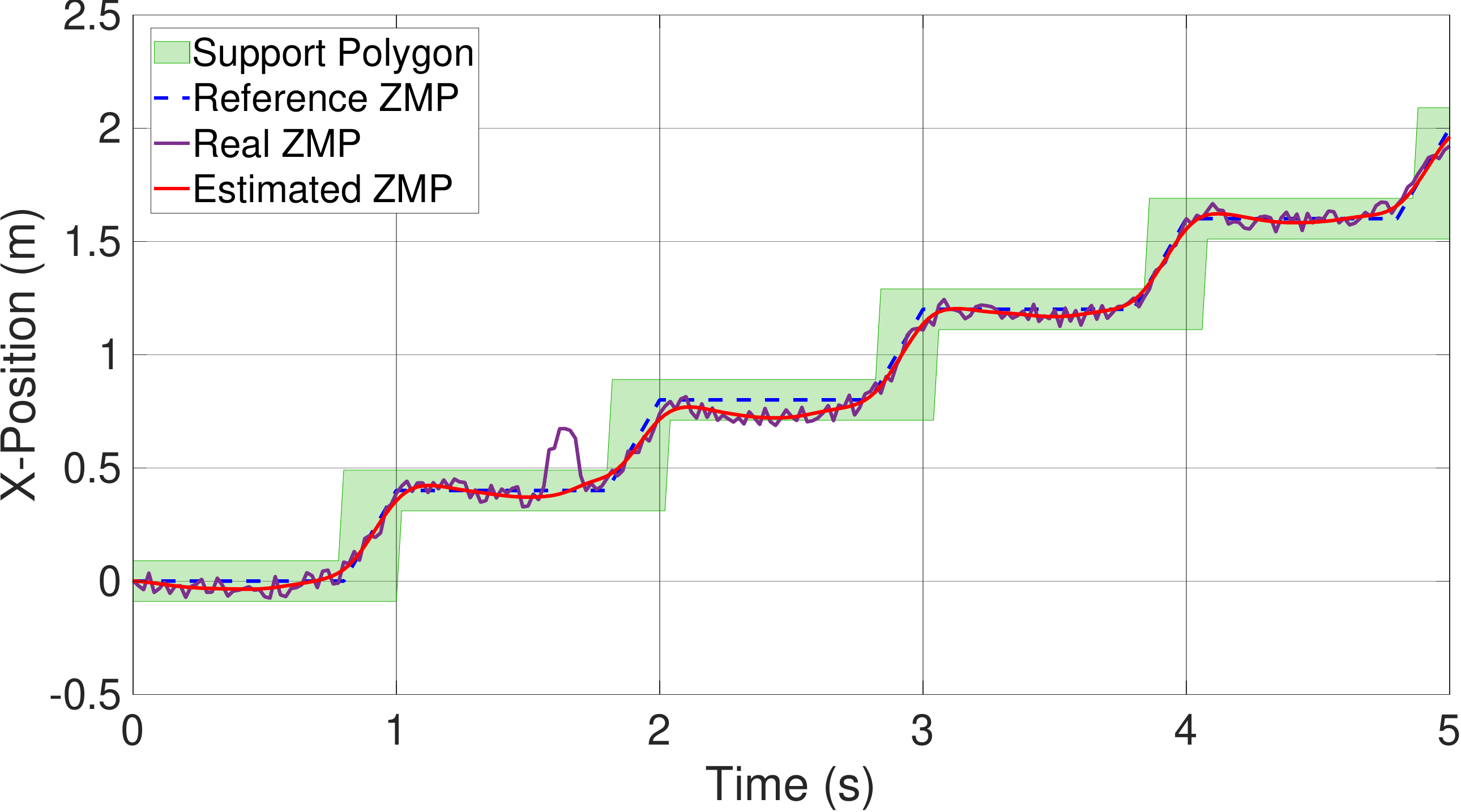}\\
		F = 300N & F = -300N\\		
		\includegraphics[width = 0.45\columnwidth, trim= 0.0cm 0.cm 0cm 0cm,clip]{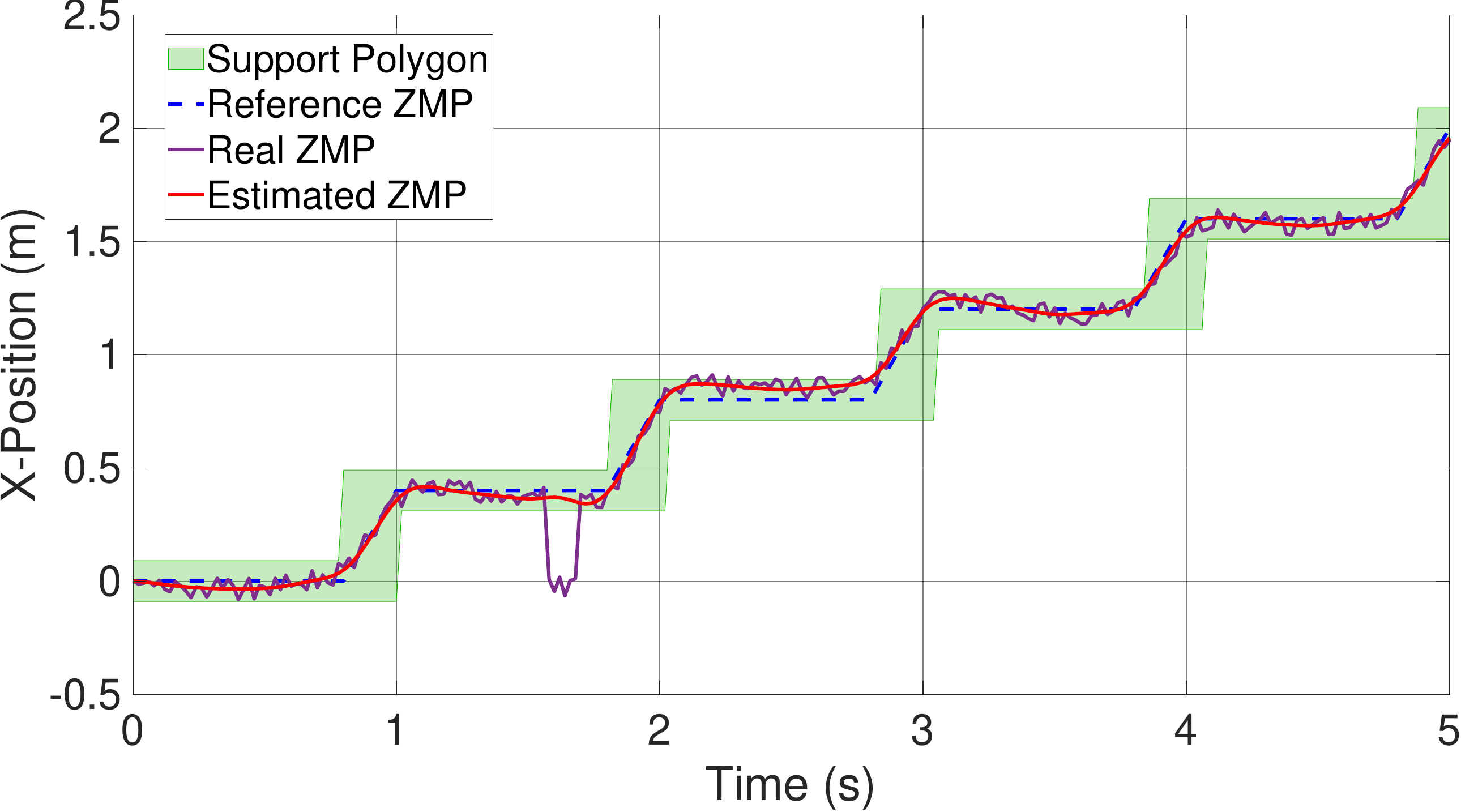}&
		\includegraphics[width = 0.45\columnwidth, trim= 0.0cm 0.cm 0cm 0cm,clip]{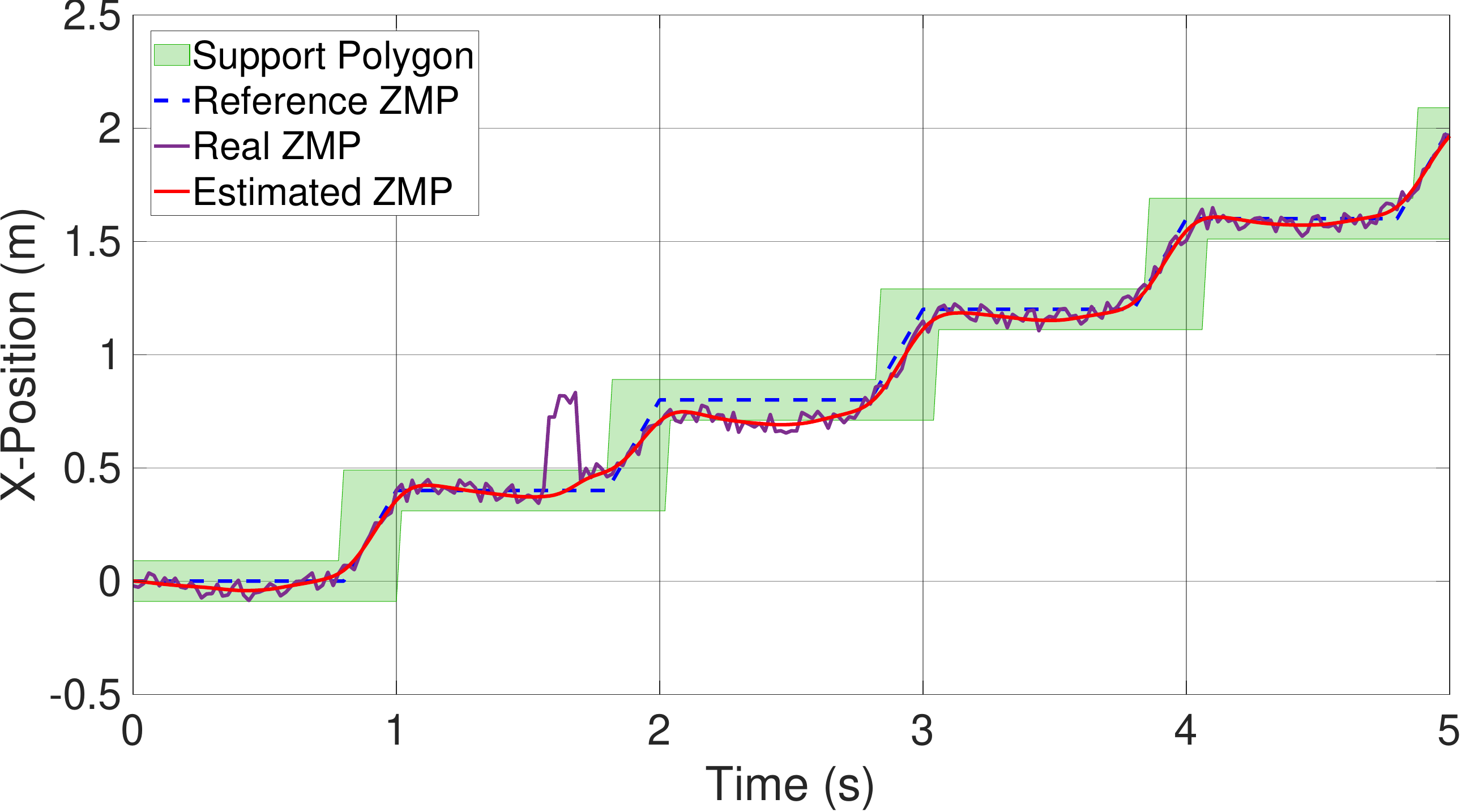}
	\end{tabular}
	\vspace{-0mm}
	\caption{ The simulation results of examining the robustness w.r.t. external disturbance: each plot represents a single simulation result. As the results show, after applying external force, ZMP~(fictitious ZMP) goes out of support polygon for a moment whose distance from the support polygon edge proportionally relates to the intensity of the perturbation.}	
	\label{fig:disturbance}
\end{figure}

\subsubsection{Robustness w.r.t. External Disturbance}
A robust controller should be able to reject an unwanted external disturbance that can occur in some situations like when a robot hits an obstacle or when it has been pushed by someone. In such situations, the controller cancels the effect of the impact and tries to keep ZMP inside the support polygon by applying compensating torque. To examine the robustness of the controller w.r.t. external disturbances, an unpredictable external force is applied to the torso of the robot while it is performing the previous scenario. The impact has been applied at $t = 1.6s$ and the impact duration is $\Delta t = 100ms$. This simulation was repeated multiple times with different amplitudes~($-300N \le F \le 300N$). Moreover, to have realistic simulations, the measurements are confounded by measurement noise~($-0.05m \le v_i\le0.05m\quad~i=1,2,3$). The simulation results are shown in Fig.~\ref{fig:disturbance}. Each plot represents the result of a single simulation. As these plots show, after applying an external force, the ZMP~(fictitious ZMP~\cite{vukobratovic2004zero}) goes out of the support polygon quickly whose distance from the support polygon edge proportionally relates to the intensity of the perturbation. The controller regains it back and keeps the stability of the robot. We increase the amplitude of the impact to find the maximum withstanding of the controller. After performing these simulations, $F= 435N$ and $F= -395N$ were the maximum levels of withstanding of the controller. According to the simulation results, the proposed controller is robust against external disturbance.
\begin{figure}[!t]
	\centering
	\begin{tabular}	{c}	
% 		\includegraphics[width = 0.75\columnwidth, trim= 0.0cm 0.cm 0cm 0cm,clip]{figures/Figures/Overall0.pdf} \\
% 		\textbf{(a)}\\
		\includegraphics[width = 0.85\columnwidth, trim= 0.0cm 0.cm 0cm 0cm,clip]{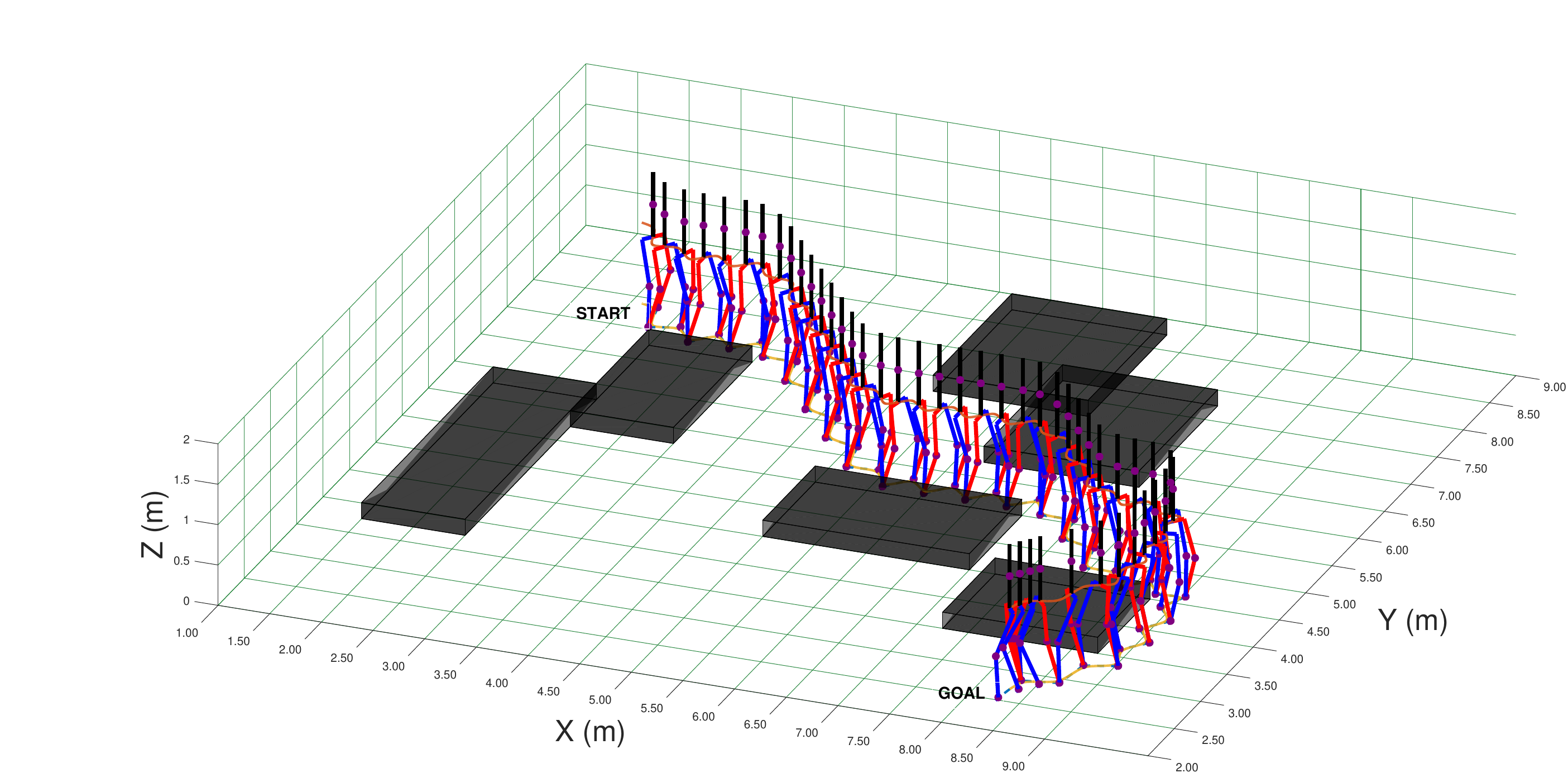} 
% 		\\
% 		\textbf{(b)}
	\end{tabular}
	\vspace{-0mm}
	\caption{ The simulation result of examining the overall performance of the proposed planner and controller. The simulated robot starts from START point and should walk towards the GOAL point while avoiding the obstacles.}
	\label{fig:sim}
\end{figure}

\subsubsection{Checking the Overall Performance}
To check the overall performance of the proposed framework, the exemplary path planning scenario which has been presented at the end of Section.~\ref{sec:ReferenceTrajectories}, is used (see Fig.~\ref{fig:PathPlanning}). The simulation results are shown in Fig.~\ref{fig:sim}. The results showed that the planner was able to generate walking reference trajectories and the controller was able to track the generated references successfully. A video of this simulation is available online at \href{https://youtu.be/zyOmgvaXyuA}{\footnotesize{https://youtu.be/zyOmgvaXyuA}}.

\subsection{Simulation using COMAN}
To validate the portability and platform-independency of the proposed framework and to show the performance of the framework in controlling a full humanoid robot, we performed a set of simulations using a simulated COMAN humanoid in the Gazebo simulator which is an open-source simulation environment developed by the Open Source Robotics Foundation (OSRF). The simulated robot is $0.95m$ tall, weighs $31kg$, and has $23$ DOF ($6$ per leg, $4$ per arm, and $3$ between the hip and torso). This robot is equipped with the usual joint position and velocity sensors, an IMU on its hip, and torque/force sensors at its ankles.  

\subsubsection{Walking Around A Disk}
\begin{figure}
	\centering
	\begin{tabular}	{c}
	\includegraphics[width=0.75\columnwidth, trim= 0.0cm 0.cm 0cm 0cm,clip]{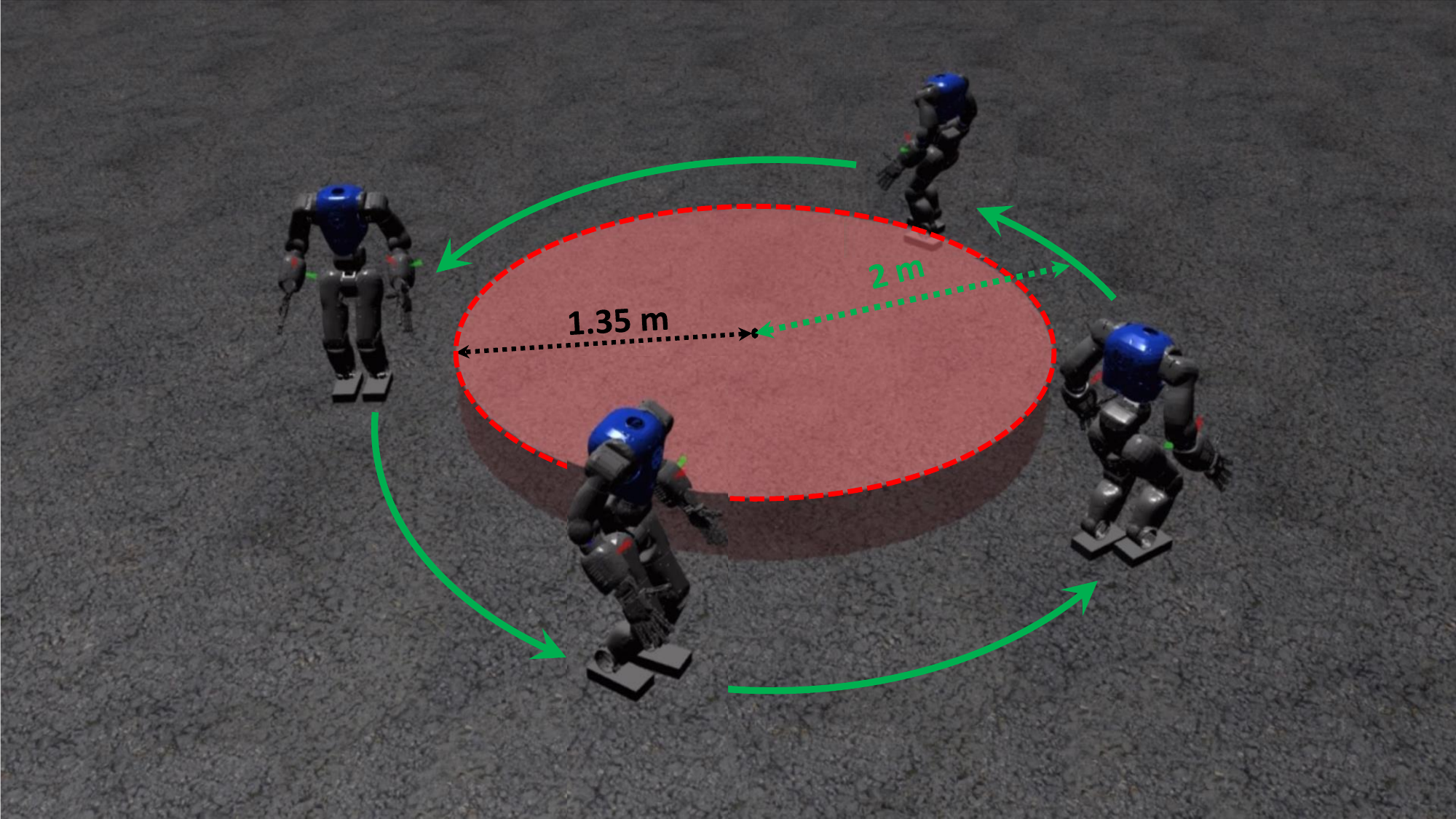} \\
	\textbf{(a)}\\
		\includegraphics[width=0.85\columnwidth, trim= 0.0cm 0.cm 0cm 0cm,clip]{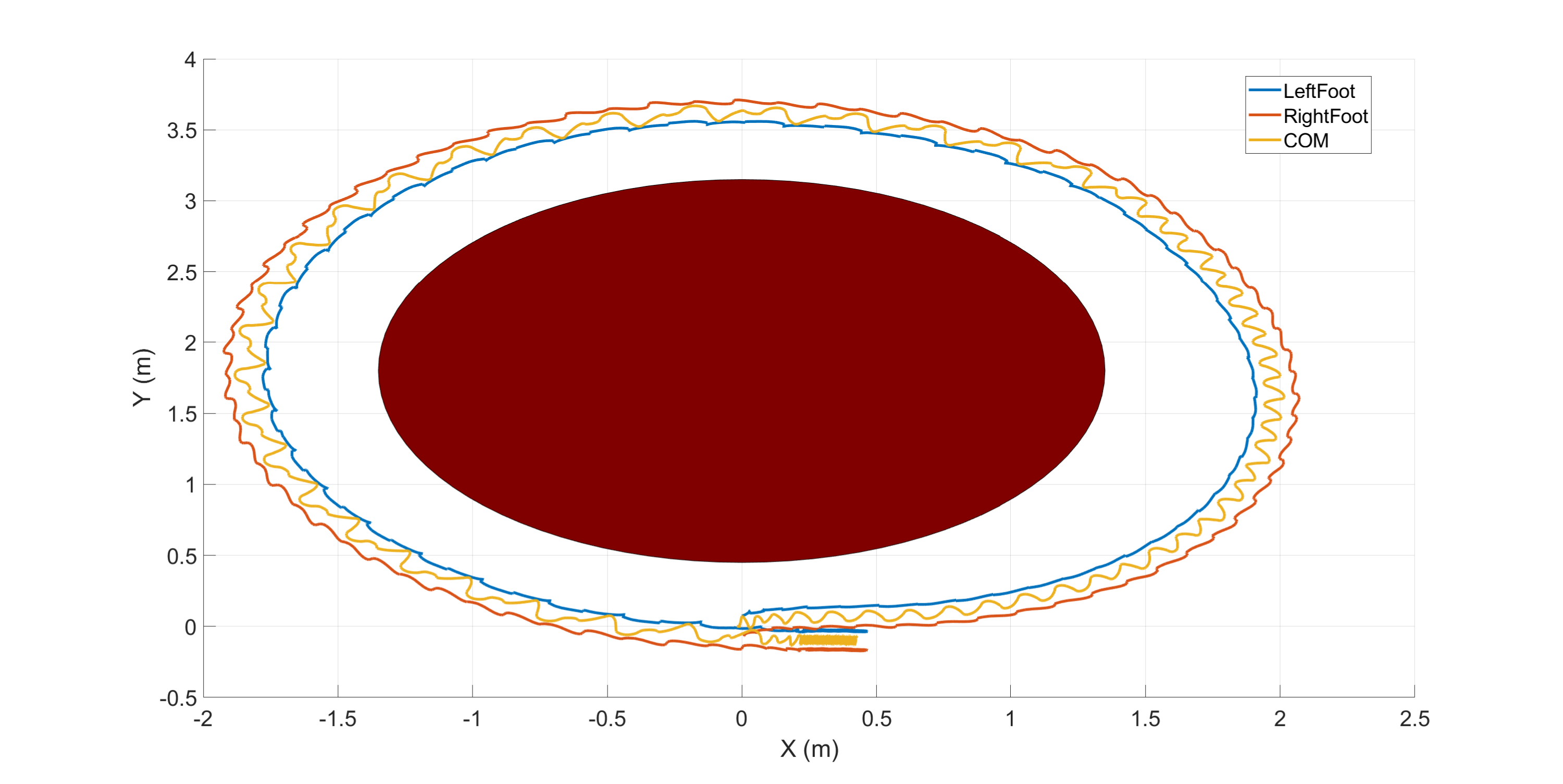} \\
		\textbf{(b)}
	\end{tabular}
	\caption{Walking around a disk: the robot should walk around the disk while keeping $2m$ distance from the center of the disk. \textbf{(a)} An overview of the scenario; \textbf{(b)} The feet and COM in the XY plane.}
	\vspace{-0mm} 
	\label{fig:walkingaround}
\end{figure}

This simulation is focused on testing the performance of the framework for combining turning and forward walking. In this simulation, the robot initially stays next to a large disk of radius $1.35m$ so that the center of the disk is $1.8m$ far from the robot. The robot is waiting to receive a start signal and once the signal is generated, the robot should walk around the disk and return to the initial point while trying to keep $2m$ distance from the center of the disk during walking. In this simulation, we fixed the step size by $0.15m$ and the turning angle will be determined based on the current position of the robot. The sequences of the experiment are shown in Fig.~\ref{fig:walkingaround}\textbf{(a)}. In this simulation, the positions of the feet and COM have been recorded and they are depicted in Fig.~\ref{fig:walkingaround}\textbf{(b)}.
The results showed that the framework is able to combine steering and forward walking command to follow a specific path. A video of this simulation is available online at \href{https://youtu.be/E8PGY05WzIQ}{\footnotesize{https://youtu.be/E8PGY05WzIQ}}.

\subsubsection{Omnidirectional Walking}

\begin{figure}[!t]
	\centering
	\begin{tabular}	{c}	
		\includegraphics[width = 0.75\columnwidth, trim= 0.0cm 0.cm 0cm 0cm,clip]{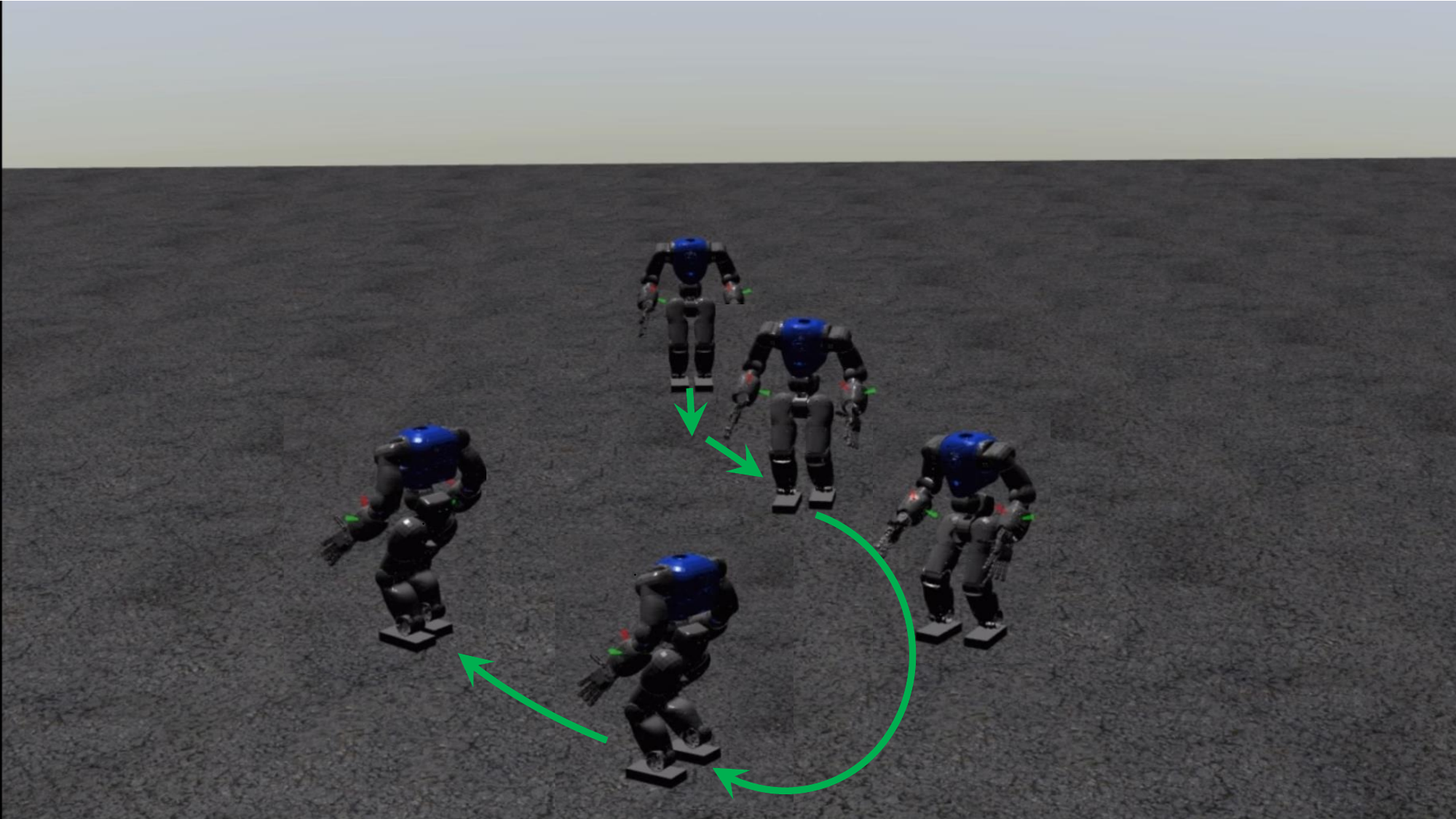} \\
		\textbf{(a)}\\
		\includegraphics[width= 0.75\columnwidth, trim= 0.0cm 0.cm 0cm 0cm,clip]{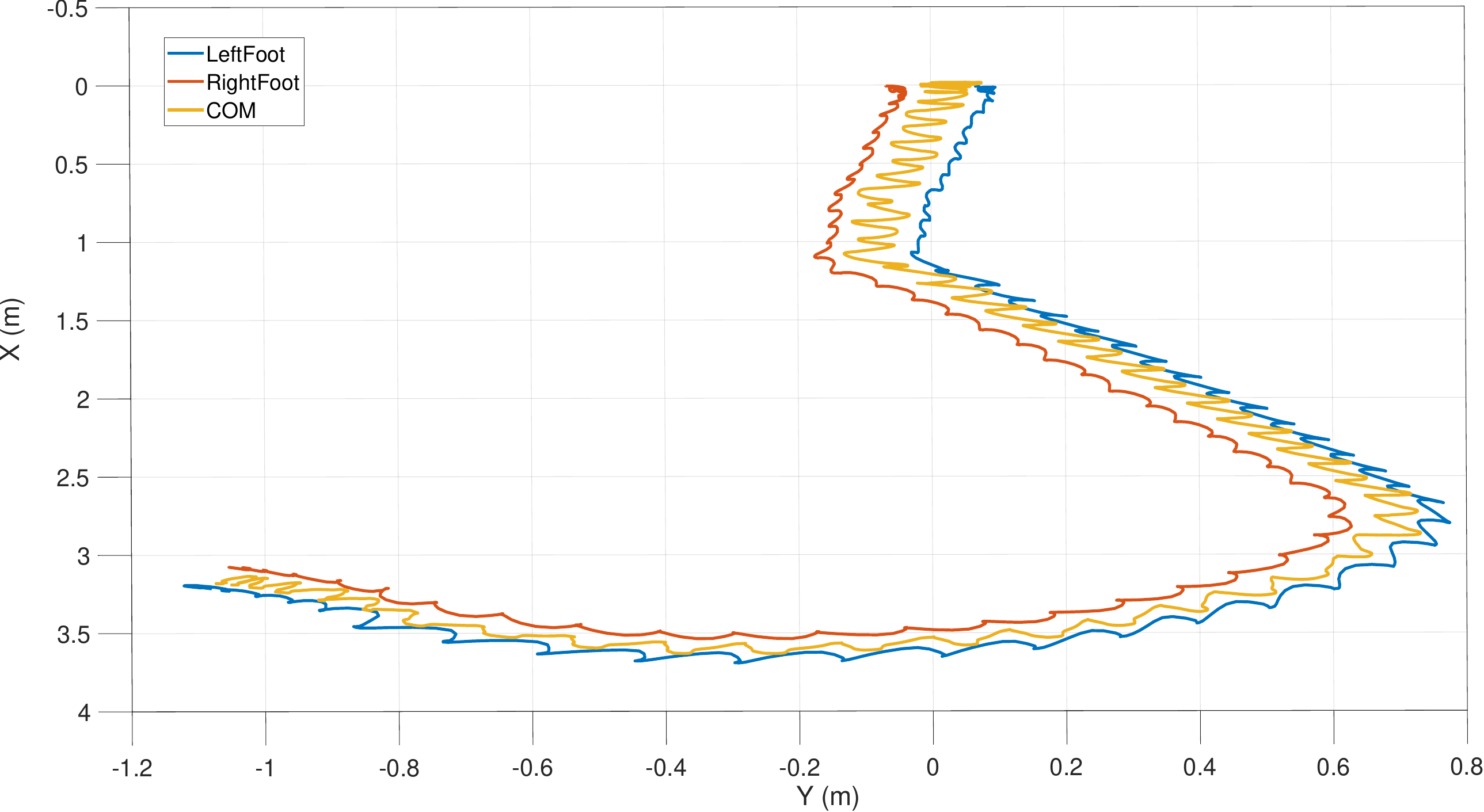} \\
		\textbf{(b)}	
	\end{tabular}
	\vspace{-0mm}
	\caption{ Omnidirectional walking scenario: \textbf{(a)} An overview of the omnidirectional walking scenario; \textbf{(b)} The feet and COM in the XY plane.}	
	\vspace{-2mm}
	\label{fig:omniwalk}
\end{figure}

This scenario is focused on validating the performance of the proposed framework for providing omnidirectional walking. In this scenario, the simulated robot should track deterministic setpoints including step length~($X$), step width~($Y$), and step angle~($\alpha$). At the beginning of the simulation, all the setpoints are zero and the robot is walking in place. At $t=2s$, the robot is commanded to walk forward~($X=0.1m, Y=0.0m, \alpha=0.0deg/s$),
At $t=10s$, the robot is commanded to walk diagonally~($X=0.10m, Y=0.05m, \alpha=0.0deg/s$), at $t=20s$, while it is performing diagonal walking, it is commanded to turn right simultaneously~($X=0.2m, Y=0.05m, \alpha=15deg/s$), finally at $t=32s$, all set points will be reset and the robot will walking in place. An overview of this scenario is depicted in Fig.~\ref{fig:omniwalk}\textbf{(a)}. During this simulation, the positions of the feet and COM have been recorded to examine the behavior of the COM while walking and it is depicted in Fig.~\ref{fig:omniwalk}\textbf{(b)}. According to the recorded data, the COM tends to the support foot during the single support phase and moves to the next support foot during the double support phase. It should be mentioned that we used a first-order lag filter to update the new setpoints to have a smooth updating. The results showed that the framework can combine all the input commands simultaneously to provide an omnidirectional walking. A video of this simulation is available online at \href{https://youtu.be/1R56m4t6cHs}{{https://youtu.be/1R56m4t6cHs}}.

\subsubsection{Push Recovery}
\begin{figure}[!t]
	\centering
	\begin{tabular}	{c}	
		\includegraphics[width = 0.75\columnwidth, trim= 0.0cm 0.cm 0cm 0cm,clip]{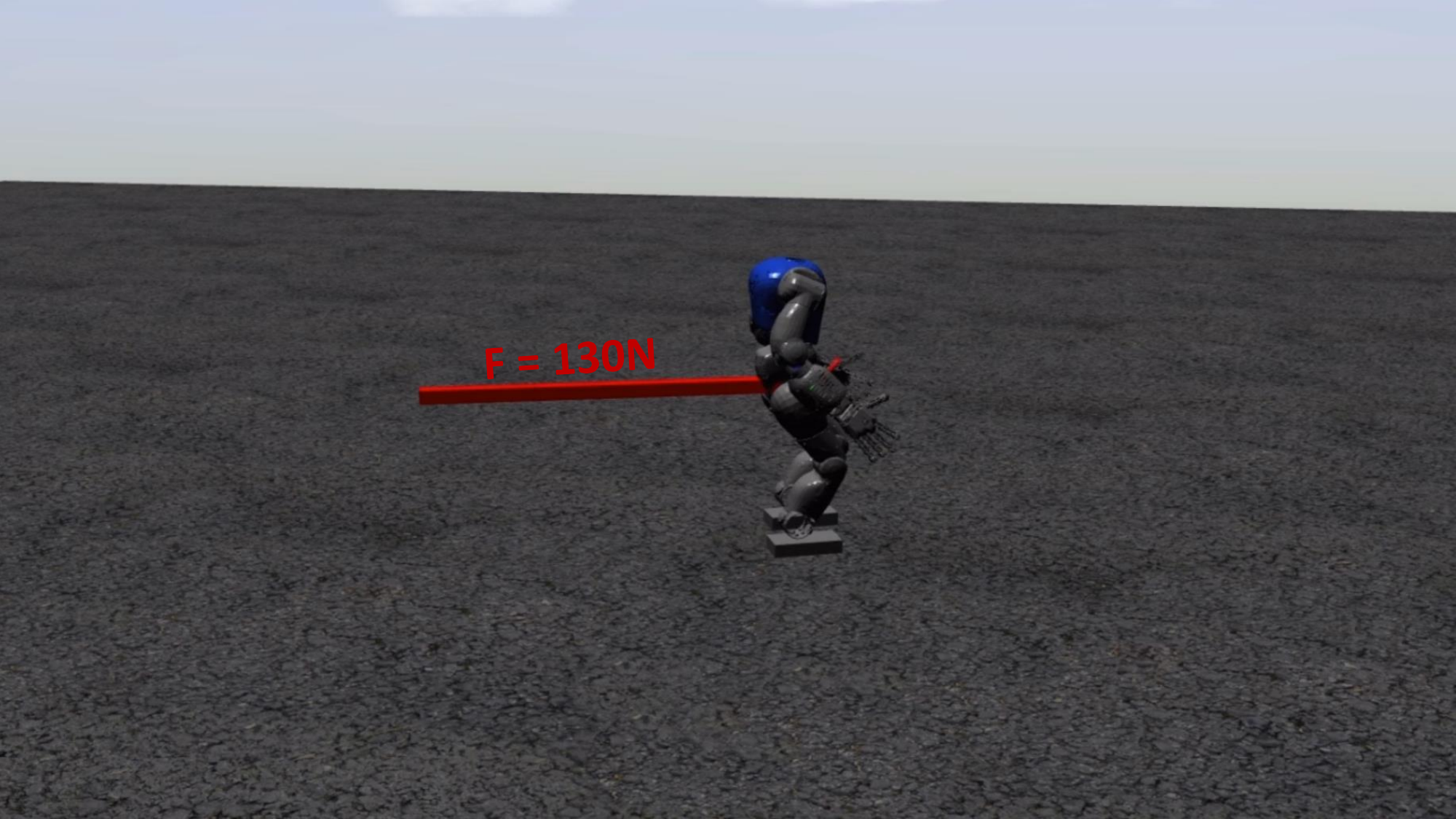} \\
		\textbf{(a)}\\
		\includegraphics[width= 0.75\columnwidth, trim= 0.0cm 0.cm 0cm 0cm,clip]{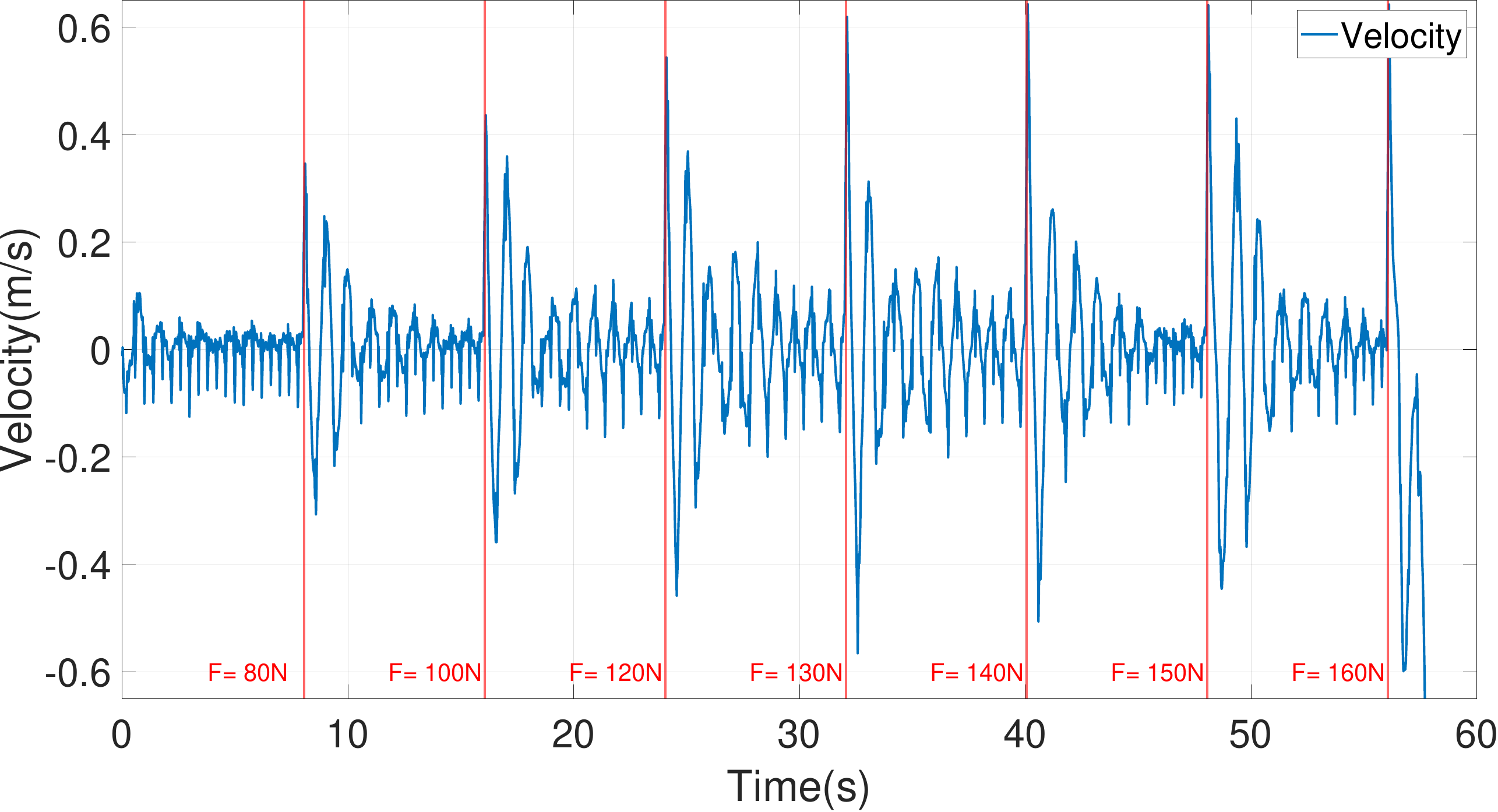} \\
		\textbf{(b)}
% 		\\
% 		\includegraphics[width= 0.85\columnwidth, trim= 0.0cm 0.cm 0cm 0cm,clip]{figures/Figures/cop_push_180N.pdf} \\
% 		\textbf{(c)}	
	\end{tabular}
	\vspace{-0mm}
	\caption{ Push recovery scenario: while the robot is walking in place, an impulsive push will be applied.  \textbf{(a)} A snapshot of push recovery scenario; \textbf{(b)} The blue line represents the velocity of COM in X direction during the simulation and the pink rectangles represent the push time and duration.}	
	\vspace{-2mm}
	\label{fig:push}
\end{figure}

The goal of this simulation is to evaluate the withstanding of the framework in terms of external disturbance rejection. In this scenario, while the robot is walking in place ($step time = 0.4s$ ), an impulsive external disturbance will be applied to the middle of its hip and the robot should reject the disturbance and keep its stability while continuing walking in place. To validate the robustness of the framework and to characterize the maximum level of withstanding of the robot, this simulation will be repeated with different amplitudes and a fixed duration of impact~($100ms$). The amplitude of the first impact is $80N$ and it will be increased every $8s$ by $20N$ while the amplitude is less than $120N$ otherwise by $10N$ until the robot falls. Fig.~\ref{fig:push}\textbf{(a)} shows a snapshot of the simulation while the robot is subject to a severe push $130N$ and capable of rejecting this disturbance and keeping its stability. The simulation results show that the framework is robust against external disturbances and $F=150N$ was the maximum level of withstanding of the robot. Fig.~\ref{fig:push}\textbf{(b)} shows the velocity of COM in X direction during this simulation. In this figure, the pink rectangles represent the push duration and as it is shown in this figure, after applying a push, the COM's velocity increases impressively and the controller could able to keep the stability. A video of this simulation is available online at \href{https://youtu.be/os3Dex07Op0}{{https://youtu.be/os3Dex07Op0}}.

\subsubsection{Walking on Uneven Terrains}
This scenario is designed to validate the performance of the framework for generating walking on uneven terrain. In this simulation, the robot is placed on an uneven terrain within a square area of side length $4m$.
\begin{figure}[!t]
	\centering
	\begin{tabular}	{c}	
		\includegraphics[width = 0.7\columnwidth, trim= 0.0cm 0.cm 0cm 0cm,clip]{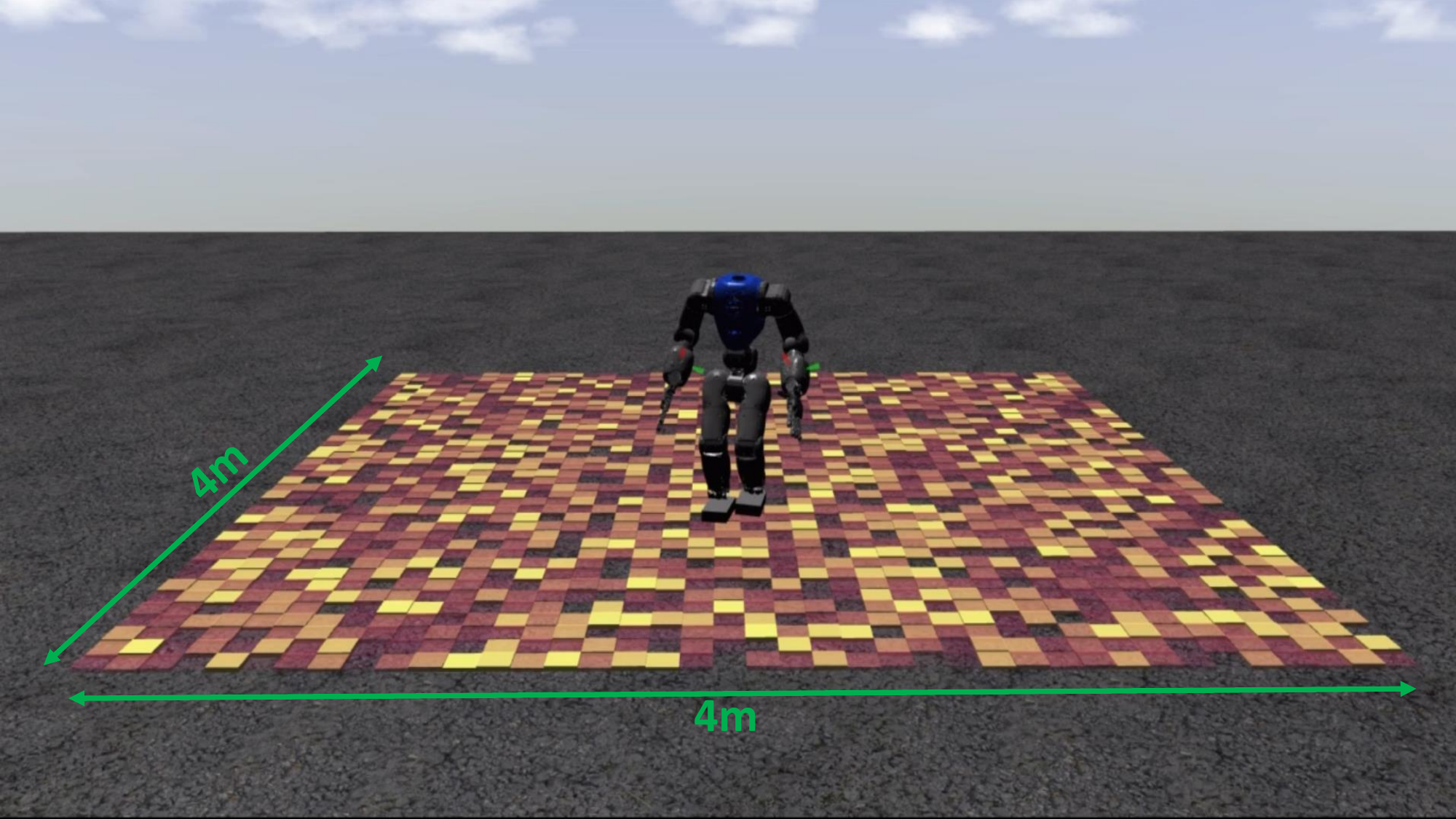} \\
		(a)\\
		\includegraphics[width = 0.7\columnwidth, trim= 0.0cm 0.cm 0cm 0cm,clip]{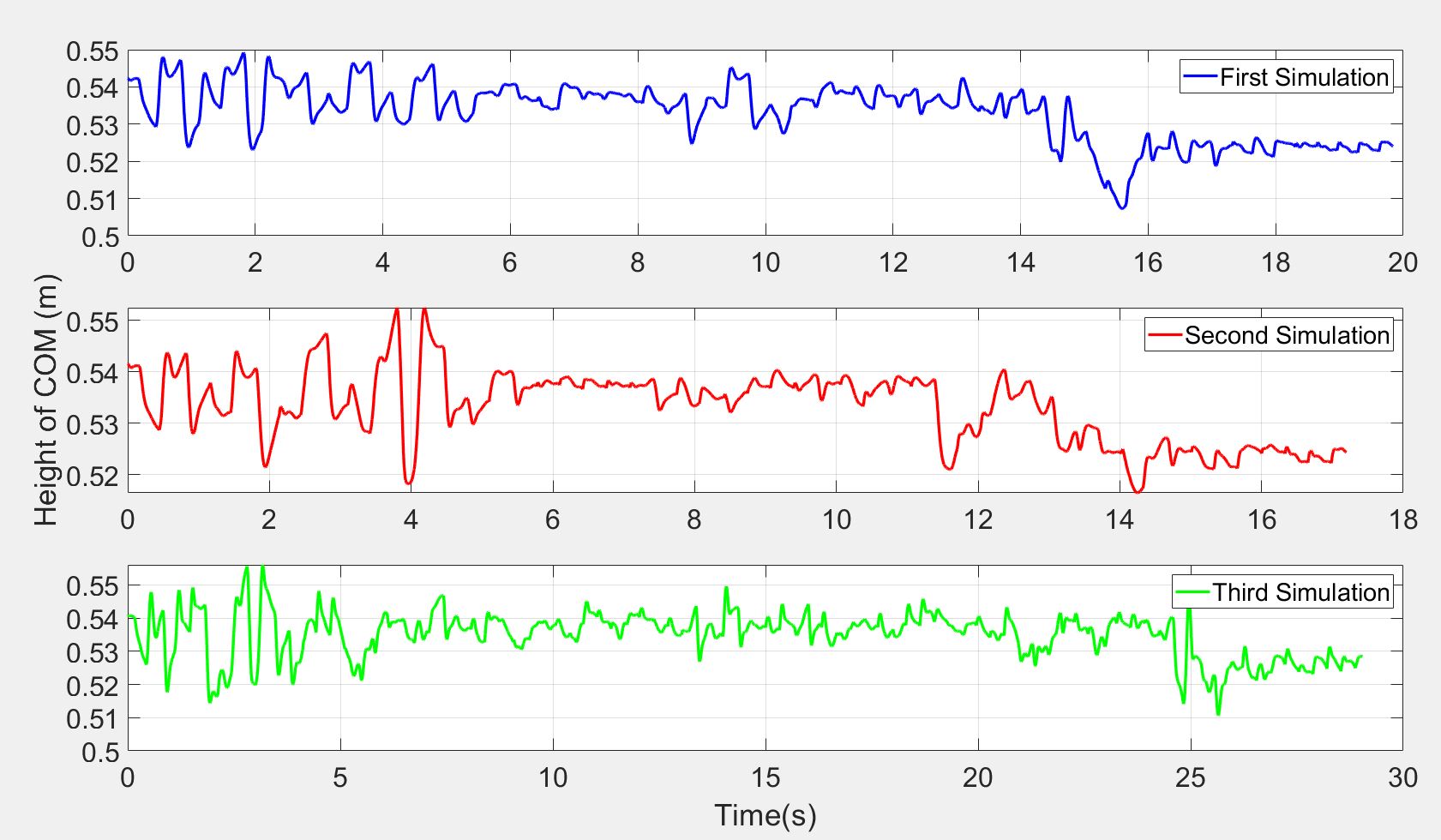} \\
		(b)	
	\end{tabular}
	\vspace{-0mm}
	\caption{ Walking on an uneven terrain: \textbf{(a)} A snapshot of the scenario; \textbf{(b)} The height variation of COM while robot is walking on the uneven terrain.}	
	\vspace{-2mm}
	\label{fig:uneven}
\end{figure}
The robot does not have any information about the terrain and should walk forward to step out of this area. In this simulation, the uneven terrain is generated by placing a set of tiles with the same size but random height (from $25mm$ to $50mm$) next to each other. To have a clear representation of the generated terrain, the color of each tile is considered as a function of its height (see Fig.~\ref{fig:uneven}\textbf{(a)}). This simulation has been repeated using three different tile's size~($100cm^2$, $400cm^2$ and $625cm^2$). The complexity of the terrain depends to the size and the height of the tiles, the terrain that is generated by the larger tile is more challenging than the others because of more ups and downs and more narrow edges which cause slipping the feet. The height variations of the COM has been recorded during the simulations and are presented in Fig.~\ref{fig:uneven}\textbf{(b)}. The simulation results showed that the framework is capable of providing stable walking on such terrains and handle such uncertainties. A video of this simulation is available online at \href{https://youtu.be/e9MK6Jy1KHg}{{https://youtu.be/e9MK6Jy1KHg}}.
\vspace{-2mm}

\section{Experiments}
\label{sec:Discussion}
The simulation results showed that the framework was able to generate robust locomotion even in challenging situations. To explore the effectiveness of the proposed framework regarding the dynamics model and the controller, we developed a baseline framework based on LIPM and our proposed framework. Then we conducted all the previous simulations using COMAN to compare the results. In \textit{Walking Around A Disk} scenario, we increased the complexity of the scenario by changing the step's lengths from $0.08m$ to $0.25m$. The results showed that the baseline framework can successfully complete the simulation by maximum step length of $0.12m$ while the proposed framework is capable of completing the simulation by $0.22m$ which is $83\%$ better than the baseline. In \textit{Omnidirectional Walking} scenario, although the current commands were challenging for the baseline, both frameworks were able to complete the simulation successfully. Therefore, we defined a scale factor to increase the complexity of the scenario by changing the input commands. The results showed that the scale factor can be increased up to $1.03$ for the baseline and $1.16$ for the proposed framework. For both frameworks, the last command~($X=0.2m, Y=0.05m, \alpha=0.26rad/s$) was the most challenging command in this simulation and the robot mostly lost its stability. Therefore, we selected this command of this simulation to compare the performance of both frameworks. The results showed that the proposed framework is $13\%$ better than the baseline. In the next simulation, to compare the maximum level of the withstanding, we performed the \textit{Push Recovery} scenario using the baseline and $F=100N$ was the maximum withstanding level which is $50\%$ less than the proposed framework. In the last simulation, we examine the capability of baseline for walking on an uneven terrain. In the two first terrains (small and medium tiles), the robot could pass almost half of the terrain and once it put its feet on a narrow edge, it lost it stability and could not be able to recover. In the third terrain, it lost its stability immediately. Therefore, the baseline could not pass successfully the uneven terrains and always fall. A summary of the simulations results is given in the Table~\ref{tb:summary}.

\begin{table}[h!]
	\centering
	\caption{Summary of the simulation results.}
	\label{tb:summary}	
		\begin{tabular}{c||c|c|c}
		\textbf{Simulation Scenario} & \textbf{Baseline framework} & \textbf{Proposed framework} & \textbf{Improvement} \\
		\hline\hline
		\textbf{Walking Around A Disk} & step length: $0.12m$  & step length: $0.22m$ & 83\% \\ \hline
		\textbf{Omnidirectional Walking} & scale factor: 1.03 & scale factor: 1.16 & 13\% \\ \hline 
		\textbf{Push Recovery} &  withstanding: 100N &  withstanding: 150N & 50\% \\ \hline 
		\textbf{Walking on Uneven Terrains} & Failed  & Success & 100\%  
		\end{tabular}
	\vspace{-4mm}
\end{table}

The simulation results validated the performance of the framework and showed that considering the dynamics of torso and legs extremely improved the performance in terms of stability and speed. Unlike~\cite{herdt2010online,brasseur2015robust} we have not used the current state of the system to adjust the planner parameters online, which improves the performance in terms of stability and speed. Also, unlike~\cite{brasseur2015robust}, we did not take into account the vertical motion of masses to keep the linearity of our dynamics model. Taking into account these motions along with the arm motions can improve the performance impressively.

\section {Conclusion}
\label{sec:CONCLUSION}
In this paper, we have developed a model-based walking framework to generate robust biped locomotion. The core of this framework is a dynamics model that abstracts the overall dynamics of a robot into three masses. In particular, this dynamics model and the ZMP concept were used to represent the overall dynamics model of a humanoid robot as a state-space system. Then, this state-space system was used to formulate the walking controller as a linear MPC which generates the control solution using an online optimization subject to a set of objectives and constraints. Later, we have presented a hierarchical planning approach which was composed of three main layers to generate walking reference trajectories. To examine the performance of the proposed planner, a path planning simulation scenario has been designed to validate the performance of the planner. Afterward, according to the planned reference trajectories, a set of numerical simulations has been performed using \mbox{MATLAB} to examine the performance and robustness of the controller. Besides, the proposed framework has been deployed on a simulated COMAN humanoid robot to conduct a set of simulations using an ODE based simulator environment. The simulation results validated the performance and robustness of the proposed framework. Additionally, the simulation results confirmed that considering the dynamics of torso and legs extremely improved the performance in terms of stability and speed.

In future work, we would like to extend this work to investigate the effect of vertical motion of the masses as well as the arm motions. Moreover, we would like to extend the framework capable of modifying online step position and duration to improve the gait stability. Additionally, we would like to develop a deep reinforcement learning module that combines with the proposed framework to regulate the framework parameters adaptively and to generate residuals to adjust the robot's target joint positions.

\section*{Acknowledgment}
This research is supported by Portuguese National Funds through Foundation for Science and Technology (FCT) through FCT scholarship SFRH/BD/118438/2016 and in the context of the project UIDB/00127/2020.

\bibliographystyle{spmpsci}   
%\bibliography{bib/refs}
\bibliography{template}
\end{document}